\pdfoutput=1

\documentclass[11pt]{article}

\usepackage[final]{acl}

\usepackage{times}
\usepackage{latexsym}

\usepackage[T1]{fontenc}

\usepackage[utf8]{inputenc}

\usepackage{microtype}

\usepackage{inconsolata}

\usepackage{amsmath}
\usepackage{amsthm}
\usepackage{amsfonts}
\usepackage{amssymb}
\usepackage{mathtools}
\usepackage{tikz}
\usepackage{mathabx}

\usepackage{comment}

\usepackage{tikz}
\usetikzlibrary{bayesnet}
\usetikzlibrary{arrows}
\usetikzlibrary{decorations}
\usepackage{subcaption}
\definecolor{ForestGreen}{RGB}{34,139,34}
\usepackage{graphicx}
\usepackage{booktabs}
\usepackage{tabularx}
\usepackage{tabulary}
\usepackage{colortbl}
\usepackage{xspace}
\newcolumntype{Y}{>{\centering\arraybackslash}X}
  \newcolumntype{P}{>{\raggedleft\arraybackslash}X}
\usepackage{multirow}
\usepackage{stfloats}
\usepackage{soul}
\usepackage{algorithm}
\usepackage[noend]{algorithmic}

\usepackage{booktabs}  %
\usepackage{array}     %
\usepackage{ragged2e}  %
\usepackage{adjustbox} %
\usepackage{setspace}  %
\usepackage{soul}

\usepackage[noabbrev,capitalise]{cleveref}

\usepackage{amsmath,amsfonts,bm}

\def\eqref#1{equation~\ref{#1}}

\def\1{\bm{1}}

\DeclareMathAlphabet{\mathsfit}{\encodingdefault}{\sfdefault}{m}{sl}
\SetMathAlphabet{\mathsfit}{bold}{\encodingdefault}{\sfdefault}{bx}{n}

\usepackage{todonotes}
\makeatletter
\newcommand*\iftodonotes{\if@todonotes@disabled\expandafter\@secondoftwo\else\expandafter\@firstoftwo\fi}

\newcommand{\icr}{ICR\textsuperscript{2}}

\definecolor{myblue}{RGB}{114,166,202}
\definecolor{myred}{RGB}{224,71,76}

\usepackage{booktabs,arydshln}
\makeatletter
\def\adl@drawiv#1#2#3{%
        \hskip.5\tabcolsep
        \xleaders#3{#2.5\@tempdimb #1{1}#2.5\@tempdimb}%
                #2\z@ plus1fil minus1fil\relax
        \hskip.5\tabcolsep}
\newcommand{\cdashlinelr}[1]{%
  \noalign{\vskip\aboverulesep
           \global\let\@dashdrawstore\adl@draw
           \global\let\adl@draw\adl@drawiv}
  \cdashline{#1}
  \noalign{\global\let\adl@draw\@dashdrawstore
           \vskip\belowrulesep}}
\makeatother

\title{Eliciting In-context Retrieval and Reasoning for 
 \\ Long-context Large Language Models}

\author{Yifu Qiu$^1$\thanks{~~Work done while the author was an intern at Apple.}, Varun Embar$^2$, Yizhe Zhang$^2$, \\ \textbf{Navdeep Jaitly$^2$, Shay B. Cohen$^1$, Benjamin Han$^2$} \\
  $^1$ University of Edinburgh, $^2$ Apple \\
 \texttt{\{yifu.qiu, scohen\}@ed.ac.uk} , \\
  \texttt{\{v\_embar, yizhe\_zhang, njaitly, ben.b.han\}@apple.com}
}

\begin{document}
\maketitle
\begin{abstract}
Recent advancements in long-context language models (LCLMs) promise to transform Retrieval-Augmented Generation (RAG) by simplifying pipelines. With their expanded context windows, LCLMs can process \textit{entire} knowledge bases and perform retrieval and reasoning directly -- a capability we define as \textbf{I}n-\textbf{C}ontext \textbf{R}etrieval and \textbf{R}easoning (\textbf{\icr{}}). However, existing benchmarks like LOFT often overestimate LCLM performance by providing overly simplified contexts. 
To address this, we introduce \icr{}, a benchmark that evaluates LCLMs in more realistic scenarios by including confounding passages retrieved with strong retrievers. We then propose three methods to enhance LCLM performance: (1) \textit{retrieve-then-generate fine-tuning}, (2) \textit{retrieval-attention-probing}, which uses attention heads to filter and de-noise long contexts during decoding, and (3) \textit{joint retrieval head training} alongside the generation head.
Our evaluation of five well-known LCLMs on LOFT and \icr{} demonstrates significant gains with our best approach applied to Mistral-7B: +17 and +15 points by Exact Match on LOFT, and +13 and +2 points on \icr{}, compared to vanilla RAG and supervised fine-tuning, respectively. It even outperforms GPT-4-Turbo on most tasks despite being a much smaller model.
\footnote{Our code and datasets are available at \url{https://github.com/apple/ml-icr2}.}
\end{abstract}

\section{Introduction}

The ability of large language models to process long contexts has significantly expanded their applicability across various domains, including book-level information retrieval \citep{dinglongrope,jin2024llm}, summarization \citep{kim2024fables,saxena-keller-2024-moviesum,qiu-etal-2023-detecting}, and question answering \citep{liu-etal-2024-lost,wang-etal-2024-leave}. They also enable more complex tasks, such as agent trajectory modeling and planning \citep{zhao2024longagent,zhang2024chain}, video captioning \citep{xue2024longvila,zhang2023simple}, and text-to-video generation \citep{wang2024loong,lin2023videodirectorgpt}. Recent advancements in long-context language models (LCLMs) hold particular promise for reshaping the Retrieval-Augmented Generation (RAG) paradigm \citep{lee2024loft,li2024alr}. With their expanded context windows, LCLMs reduce reliance on complex pipelines required by context-length limitations and simplify knowledge updates by allowing context modifications. For instance, LCLMs can accommodate entire knowledge bases within the context windows, effectively serving as working memory for new queries.

Achieving this goal requires LCLMs to effectively retrieve and reason within their ``contextual knowledge base,'' a capability we define as \textbf{I}n-\textbf{C}ontext \textbf{R}etrieval and \textbf{R}easoning (\textbf{\icr{}}). However, existing benchmarks often fail to accurately evaluate this capability. For example, Needle-in-a-Haystack (NIAH; \citealt{niah_github}) is a popular test to determine whether a model can retrieve a ``needle'' (a specific fact or statement) randomly inserted into a ``haystack'' (a corpus). Yet, the semantic discontinuity between the needle and the haystack can unintentionally reveal the needle's location, making the task overly simple. LOFT \citep{lee2024loft}, the first large-scale benchmark for evaluating retrieval and reasoning within contextual knowledge bases, uses human-annotated relevant documents as the needle, and fills the haystack with randomly sampled documents from an external knowledge base. However, this random sampling results in a context with virtually no \textit{confounding} information that is relevant but misleading, causing LOFT to significantly overestimate LCLM performance.

To bridge this evaluation gap, we introduce \icr{}, a novel and challenging benchmark designed to assess LCLMs under more realistic conditions. \icr{} builds upon KILT \citep{petroni-etal-2021-kilt}, a comprehensive knowledge base sourced from Wikipedia. Unlike LOFT, which relies on random sampling, \icr{} uses strong retrievers to select challenging confounding documents, creating a more difficult ``haystack.'' Experimental results reveal that current LCLMs struggle on \icr{}, with exact match rates dropping by up to 51\% compared to evaluations on LOFT. These findings underscore the significant challenges LCLMs face in accurately performing in-context retrieval in realistic scenarios.

We next explore improving LCLMs' in-context retrieval and reasoning capabilities. While RAG demonstrates strong results, it remains hindered by complex multi-stage pipelines. Encouraged by recent studies such as LLM2Vec \citep{behnamghader2024llm2vec, ma2024fine} showing that LCLMs can be effectively adapted for accurate retrieval tasks, offering a more natural approach by enabling joint optimization of both retrieval and generation steps, we propose three approaches: (1) \textbf{Retrieve-then-generate fine-tuning}: Inspired by the distillation of step-by-step reasoning abilities \citep{shridhar2023distilling, hsieh-etal-2023-distilling}, we train LCLMs to frame tasks as two-hop reasoning chains. In this formulation, LCLMs first retrieve relevant information from the context and then generate the final responses. (2) \textbf{Retrieval-attention probing}: At inference time, we probe attention heads activated for in-context retrieval \citep{wu2024retrievalHead} and use their top predictions to filter out confounders from lengthy contexts. (3) \textbf{Joint retrieval head training}: We introduce an architectural modification to equip LCLMs with a dedicated retrieval head, enabling joint optimization of retrieval and generation.

We conduct extensive experiments using five LCLMs on both LOFT and our \icr{} benchmark. We compare our methods against baselines including Vanilla RAG, Closed-book, Oracle RAG, and supervised fine-tuning (SFT). Notably, our best approach, applied to Mistral-7B with a 32K token limit, has the best performance across the tasks. It outperforms Vanilla RAG and SFT baselines by an average of +17 and +15 points measured by Exact Match on LOFT, and by +13 and +2 points on \icr{}, respectively. The approach achieves performance comparable to the state-of-the-art GPT-4, despite using only 7B parameters. We also provide in-depth analyses of our approaches.

Our key contributions are as follows:
\begin{itemize}
    \item We introduce \icr{}, a realistic and challenging benchmark for evaluating the in-context retrieval and reasoning capabilities of long-context language models (LCLMs), demonstrating that existing benchmarks overestimate model performance.
    \item We explore three novel methods to enhance LCLMs ranging from supervised fine-tuning, inference-time approach, to model architecture modification.
    \item Our best approach, applied to a small LCLM, tightens the performance gap with the Oracle RAG while beating the other baselines. It even matches GPT-4 on both in-domain (\icr{}) and out-of-domain benchmarks (LOFT), albeit with a much smaller model size.
\end{itemize}

\section{Related Work}

Long-context large language models (LCLMs) have garnered significant attention for their ability to process extended sequences. Models like Longformer \citep{Beltagy2020Longformer} and BigBird \citep{zaheer2020big} introduced sparse attention mechanisms to efficiently handle long context. Scaling efforts, exemplified by GPT-4 \citep{achiam2023gpt}, underscore the importance of expanding context windows for the long-context tasks. 
Data engineering methods \citep{fu-data-engineering-llm, xiong2024effective, jin2024llm}, expanding positional encoding \citep{dinglongrope}, and the parameter-efficient fine-tuning \citep{chenlonglora} have been effective.

LCLMs hold the promise in reshaping RAG (\citealt{achiam2023gpt,jiang2023mistral,yang2024qwen2,abdin2024phi}). By replacing static knowledge bases with the contextual one, this new paradigm simplifies the deployment by discarding intermediate components while enabling the direct updates on LCLM's knowledge \citep{lee2024loft}. 

Unlike the existing works, we systematically evaluate the use of LCLMs for scenarios where knowledge bases are directly placed as the context with a novel benchmark, \icr{}. \icr{} is different with the popular long-context benchmark such as \citep{bai-etal-2024-longbench} by emphasizing in the contextual confounders.
We also show that with targeted enhancements, small-scale LCLMs can achieve performance comparable to state-of-the-art models.

\section{Are LCLMs Competent for RAG?}

\subsection{LLMs Are Sensitive to the Confounders}

\begin{figure*}[h]
    \centering
    \includegraphics[width=0.8\linewidth]{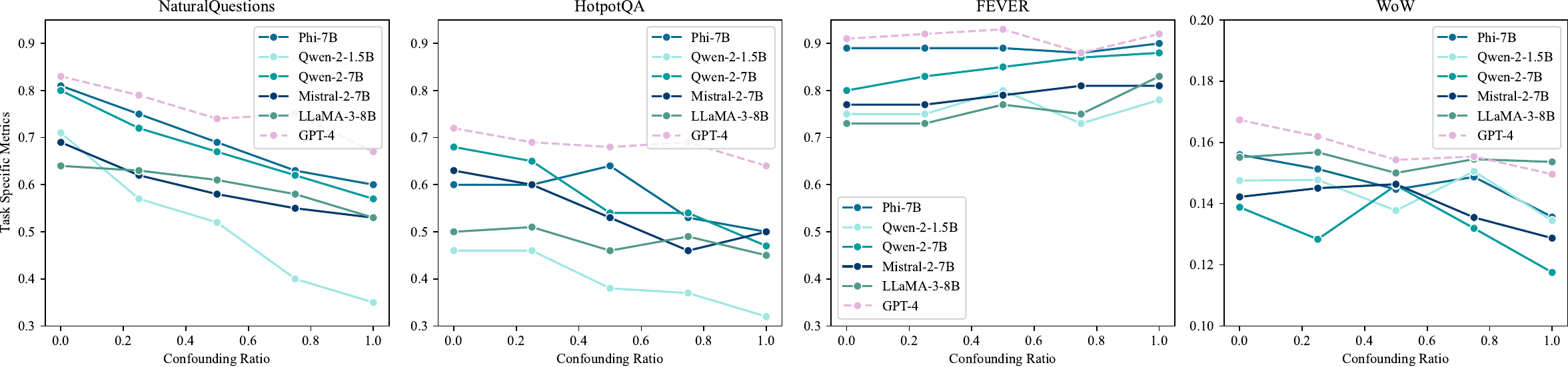}
    \caption{Task-specific performance of five LCLMs on 32K \icr{} test sets with varying confounding ratios. For all tasks, a higher value of task-specific metric indicates a better performance. 
    }
    \label{fig:varied-conf-ratio}
    \vspace{-0.5cm}
\end{figure*}

LOFT \citep{lee2024loft} introduces a Corpus-in-Context (CiC) approach for retrieval-augmented generation, integrating a large-scale external knowledge base directly into the LLM's context. However, LOFT operates under the assumption that the context is free of \textit{confounders}. In practice, a corpus often contains confounders -- documents related to the query but potentially leading to incorrect answers. For example, given the query “\textit{Who is the 44th U.S. President?}”, the corpus might include documents about the other presidents, such as “\textit{The 45th U.S. President is Donald Trump}”, which could mislead the LLM. Consequently, LOFT’s design reduces the complexities of using real-world corpora for RAG, potentially overestimating the performance of LCLMs. 

We show that LCLMs are indeed sensitive to the confounders missed in LOFT. We construct multiple test sets with varying \textit{confounding ratios} -- $\{0, 25\%, 50\%, 75\%, 100\%\}$ -- and evaluate the LCLMs under zero-shot settings.
The confounding ratio $p$ denotes the proportion of confounding context that are selected by retrievers while filtering out the gold provenance, with the remaining, i.e., $(100\%-p)$, randomly sampled from an external knowledge base.
At $p=0$, all confounding passages in the contextual knowledge base are randomly sampled, equivalent to the setup in LOFT \citep{lee2024loft}. Conversely, at $p=1$, all confounders are selected by the retrievers.

We evaluate five LCLMs with a context length of at least 32K tokens: Phi-3-7B \citep{abdin2024phi}, Qwen-2-1.5B/7B \citep{yang2024qwen2}, Mistral-003-7B \citep{jiang2023mistral}, LLaMA-3-Instruct-8B \citep{dubey2024llama-3-technical-report} and GPT-4-Turbo \citep{achiam2023gpt}. Our findings in Figure~\ref{fig:varied-conf-ratio}, demonstrates that LCLM performance is highly sensitive to confounders, with performance generally degrading as the confounding ratio increases. This indicates that confounders in \( P^- \) obtained via retrievers, as in \icr{}, pose greater challenges for the models compared to those randomly sampled in LOFT. These retriever-selected passages are relevant to the queries but fail to lead to the correct answers, making them particularly difficult for LCLMs to handle. Consequently, benchmarks like LOFT that overlook these “strong” confounders risk overestimating model performance, as confounders are prevalent in real-world scenarios.

\subsection{\icr{} Benchmark}
\label{sec:icr2-benchmark}

\begin{figure}
    \centering
    \includegraphics[width=\linewidth]{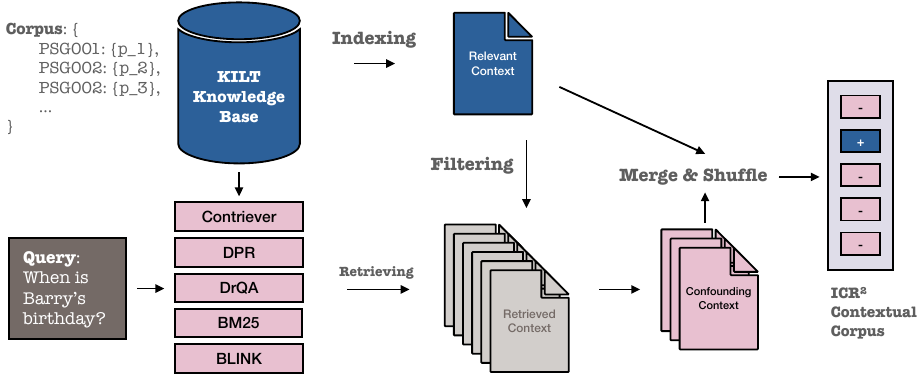}
    \caption{Construction pipeline for \icr{}. Based on KILT, we use five strong retrievers to retrieve confounders to serve as the “haystack”, while the human-annotated contexts are used as the “needle”.}
    \label{fig:icr2-creation}
    \vspace{-10pt}
\end{figure}

To address this limitation, we propose an alternative benchmark, \icr{}, which leverages strong retrievers to identify and incorporate these confounders into the contextual corpus used in CiC, providing a more realistic and challenging evaluation framework.
We choose KILT \citep{petroni-etal-2021-kilt}, a comprehensive suite of benchmarks designed for knowledge-intensive NLP tasks, to be our external knowledge base (Figure~\ref{fig:icr2-creation}). KILT covers tasks such as question answering \citep{kwiatkowski-etal-2019-natural,yang2018hotpotqa}, fact verification \citep{thorne2018fever}, and dialogue completion \citep{dinanwizard}, all paired with a single  Wikipedia snapshot. Each KILT instance $ \langle q, a, P^+ \rangle$ consists of a query $q$, the reference answer $a$, and its provenances $P^+ = \{p_1, \dots, p_m\}$, which is a set of relevant Wikipedia pages and specific locations within them that support the answer.

Building on KILT, each \icr{} instance is represented as $\langle q, a, C \rangle$, where $q$ and $a$ is a query-answer pair from KILT, and $C$ is the contextual knowledge base required to answer the query. At test time, a standardized prompt template, ${\texttt{rag\_prompt}}[C]$, is applied to the context, and the result is sent to an LCLM to generate the final answer for evaluation.
To construct $C$, we first include all provenances $P^+$ from KILT as the positive passages. For confounder $P^-$, unlike LOFT's random sampling approach, we run five strong retrievers on queries formed by concatenating $q$ with $a$ to select Top-$K$ results as the confounding passages to be included in $C$.

\begin{table}[!t]
\centering
\begin{adjustbox}{width=\linewidth}
{\small
\begin{tabular}{@{}p{\linewidth}@{}}
\toprule
\textbf{Query:} \\ 
Where are the giant redwoods located in California? \\ 
\midrule

\textbf{Reference Answer:} \\ 
Humboldt County \\ 
\midrule

\textbf{Contextual Knowledge Base:} \\
\texttt{ID: PSG001} \\ 
\textit{Title: 2010–11 California Golden Bears Men's Basketball Team} \\ 
\textbf{Context:} 
{ 
The 2010–11 \textcolor{myred}{California Golden Bears men's basketball team represented the University of California, Berkeley} in the 2010–11 NCAA Division I men's basketball season. } \\[0.5em]

\texttt{ID: PSG002} \\ 
\textit{Title: Redwood National and State Parks} \\ 
\textbf{Context:} 
{ 
Scenes set on the forest moon Endor in Star Wars were filmed in \textcolor{myblue}{the Tall Trees Redwood Grove in the northern part of Humboldt County}, though the majority of filming was in private and public forests near the town of Smith River ... } \\[0.5em]

... [90 context are omitted for the simplicity] ... \\ [0.5em]

\texttt{ID: PSG100} \\ 
\textit{Title: Malibu Grand Prix} \\ 
\textbf{Context:} 
{ 
... and San Antonio, Texas. Palace operates additional locations in Los Angeles, California and Dallas, Texas. \textcolor{myred}{The Redwood City, CA location closed on August 18, 2013 and the San Antonio location closed on September 7, 2015.}} \\
\bottomrule
\end{tabular}
}
\end{adjustbox}
\caption{An example \icr{} instance. We show the query with the reference answer, and the constructed contextual knowledge base. We highlight the true provenance in \textcolor{myblue}{blue} and the confounding passages in \textcolor{myred}{red}.}
\label{tab:data-instance}
\end{table}

We show the pipeline for \icr{} creation in Figure~\ref{fig:icr2-creation}.
To eliminate bias in the retrievers' preferences during the construction process, we pool the Top-$K$ predictions from all retrievers and uniformly sample passages from top to bottom until reaching the maximum allowable size for the contextual knowledge base. The retrievers used include Contriever \citep{izacard-contriever}, DPR \citep{karpukhin2020DPR}, DrQA \citep{chen2017reading-drqa}, BM25 \citep{robertson1976relevance-bm25,robertson2009probabilistic-bm25}, and BLINK \citep{wu2019-BLINK}. For passage-level retrievers such as Contriever, BM25, and DPR, as input we chunk all documents in KILT's knowledge base and exclude passages overlapping with any annotated provenance in $P^+$. We also set $K=200$ for the 32K version of \icr{}. For document-level retrievers like DrQA and BLINK, we set $K=20$, and chunk retrieved documents using the same process as above. For all retrievers, we exclude retrieved passages overlapping with $P^+$ or containing exact substring matches with $a$, and the remaining passages are collected to form the confounding provenance set, $P^-$. The contextual corpus $C$ is then constructed by concatenating $P^+$ with $P^-$ up to the maximum length, i.e., $C = [P^+; P^-]$. Finally, we shuffle all items in $C$ to remove position bias. 

Table~\ref{tab:benchmark-statistics} compares statistics between LOFT and our benchmark \icr{}.
An example of an \icr{} instance is provided in Table~\ref{tab:data-instance}, and the \texttt{rag\_prompt} template used is detailed in Table~\ref{tab:rag-prompt-templates} in Appendix~\ref{appendix:rag-prompt-templates}.

\subsection{State of LCLMs Performance on \icr}
\label{sec:lclm-performance-icr}

\begin{table*}[t]
\centering
\resizebox{0.85\linewidth}{!}{%
\begin{tabular}{lcccccccccc}
\toprule
\small
 & \multicolumn{4}{c}{\textbf{LOFT}} & \multicolumn{5}{c}{\textbf{\icr}} & \multirow{2}{*}{\textbf{All}} \\ 
\cmidrule(lr){2-5} \cmidrule(lr){6-10}
 & NQ & HotpotQA & MUSIQUE & Avg & NQ & HotpotQA & FEVER & WoW & Avg &  \\ 
\midrule
\small \textbf{Closed-book} & & & & & & & & & & \\
Qwen-2-1.5B-32K & 0.21 & 0.22 & 0.03 & 0.15 & 0.29 & 0.24 & 0.62 & 0.15 & 0.32 & 0.25 \\
Qwen-2-7B-32K & 0.31 & 0.30 & 0.07 & 0.23 & 0.35 & 0.30 & 0.71 & 0.15 & 0.38 & 0.31 \\
Mistral-2-7B-32K & 0.46 & 0.38 & 0.05 & 0.30 & 0.49 & 0.38 & 0.73 & 0.14 & 0.43 & 0.38 \\
Phi-3-7B-128K & 0.39 & 0.35 & 0.09 & 0.28 & 0.46 & 0.32 & 0.73 & 0.13 & 0.41 & 0.35 \\
LLaMA-3-instruct-8B & 0.41 & 0.19 & 0.01 & 0.20 & 0.44 & 0.19 & 0.53 & 0.14 & 0.32 & 0.27 \\
GPT-4-turbo & \textbf{0.58} & \textbf{0.57} & \textbf{0.28} & \textbf{0.48} & \textbf{0.61} & \textbf{0.50} & \textbf{0.85} & \textbf{0.16} & \textbf{0.53} & \textbf{0.51} \\
\midrule
\small \textbf{Vanilla RAG} & & & & & & & & & & \\
Qwen-2-1.5B-32K & 0.61 & 0.43 & 0.12 & 0.39 & 0.35 & 0.32 & 0.78 & 0.12 & 0.39 & 0.39 \\
Qwen-2-7B-32K & 0.79 & 0.61 & 0.29 & 0.56 & 0.57 & 0.47 & 0.88 & 0.13 & 0.51 & 0.53 \\
Mistral-2-7B-32K & 0.64 & 0.62 & 0.27 & 0.51 & 0.53 & 0.50 & 0.81 & 0.13 & 0.49 & 0.50 \\
Phi-3-7B-128K & 0.76 & 0.68 & 0.41 & 0.62 & 0.60 & 0.50 & 0.90 & 0.14 & 0.53 & 0.57 \\
LLaMA-3-instruct-8B & 0.45 & 0.56 & 0.16 & 0.39 & 0.53 & 0.45 & 0.83 & \textbf{0.15} & 0.49 & 0.45 \\
GPT-4-turbo & \textbf{0.85} & \textbf{0.78} & \textbf{0.51} & \textbf{0.71} & \textbf{0.67} & \textbf{0.64} & \textbf{0.92} & \textbf{0.15} & \textbf{0.59} & \textbf{0.65} \\
\midrule
\small \textbf{Oracle RAG} & & & & & & & & & & \\
Qwen-2-1.5B-32K & 0.80 & 0.76 & 0.44 & 0.67 & 0.77 & 0.65 & 0.86 & 0.15 & 0.61 & 0.63 \\
Qwen-2-7B-32K & 0.88 & 0.81 & 0.56 & 0.75 & 0.83 & 0.80 & 0.91 & 0.17 & 0.68 & 0.71 \\
Mistral-2-7B-32K & 0.89 & 0.81 & 0.41 & 0.70 & 0.83 & 0.81 & \textbf{0.94} & 0.18 & \textbf{0.69} & 0.70 \\
Phi-3-7B-128K & \textbf{0.90} & 0.85 & 0.63 & 0.79 & \textbf{0.87} & 0.81 & 0.91 & 0.14 & 0.68 & 0.73 \\
LLaMA-3-instruct-8B & 0.81 & 0.77 & 0.60 & 0.73 & 0.80 & 0.74 & 0.86 & \textbf{0.19} & 0.65 & 0.68 \\
GPT-4-turbo & 0.88 & \textbf{0.87} & \textbf{0.72} & 0.82 & 0.79 & \textbf{0.82} & \textbf{0.94} & 0.18 & 0.68 & \textbf{0.74} \\
\bottomrule
\end{tabular}}
\caption{Performance evaluation of six LCLMs on LOFT and \icr{} benchmarks. All models benefit from the Vanilla RAG approach, but a gap remains between vanilla and oracle RAG performance.}
\label{tab:icr2-lclm-perf}
\vspace{-10pt}
\end{table*}

We evaluate five LCLMs that support 32K or longer context length using LOFT \citep{lee2024loft} and \icr{} benchmarks, with each one tested with input up to 32K tokens. 
We include three baselines for comparison: (1) \textbf{Vanilla RAG}, where the CiC prompt from \citet{lee2024loft} is used; (2) \textbf{Closed-book}, where the entire context ($C$) is removed to evaluate LCLM performance based solely on parametric knowledge; and (3) \textbf{Oracle RAG}, which includes only the ground-truth relevant contexts $Z^*$ in $C$ to estimate an upper-bound performance.

Table~\ref{tab:icr2-lclm-perf} presents our results. Overall, we observe that the Vanilla RAG significantly outperforms the closed-book baseline for most models, except for Qwen-2-1.5B on the WoW dataset. This trend persists even with the state-of-the-art GPT-4-Turbo, suggesting that external knowledge remains crucial for knowledge-intensive tasks despite the large parametric capacity of LCLMs. Furthermore, RAG performance on \icr{} lags notably behind that on LOFT, reflecting \icr’s greater complexity due to its more realistic context construction. Lastly, the large performance gap between RAG and Oracle setups on both benchmarks underscores the impact of confounding information, which hampers accurate retrieval and response generation in LCLMs.

\section{Eliciting \icr{} for LLMs}
\label{sec:eliciting_icr}

\subsection{Retrieve-then-generate Fine-tuning}\label{sec:retrieve-then-generate-tuning}

Compared to the standard supervised fine-tuning where a model is asked to generate a \textbf{D}irect \textbf{A}nswer (\textbf{DA}) to a given query, our first proposal, \textit{retrieve-then-generate}, is a two-step process: the model first retrieves relevant passages from context and then generates the final answer based on the context and retrieved passages, all in one decoding pass. Formally, we train an LCLM to optimize,
\begin{equation}
\label{equ:retrieve-then-generate_equ}
    p(y \mid q, c) = \sum_{z_i \in Z} p(y \mid q, c, z_i)p(z_i \mid q, c),
\end{equation}
where $q,y$ is the query and target, respectively, $c$ is the contextual knowledge base, and $Z$ is the collection of all relevant passages necessary for answering $q$. This objective can be easily integrated into the next-token prediction task trained with the maximum likelihood estimation, just by sequentially executing the retrieval and generation in a single forward pass. The overall loss is given as
\begin{equation}
   \mathcal{L} = \frac{1}{N} \sum_{i=1}^{N} \left( \log p(Z_i^* | q_i, c_i) + \log p(y_i^* | Z_i^*, q_i, c_i)\right),
\end{equation}
where $N$ is the number of training samples, and $Z^*_i=\{ z^*_{i,1}, z^*_{i,2}, \dots, z^*_{i,|Z^*_i|}\}$ is the collection of all relevant passages for the query $q_i$ and contextual knowledge base $c_i$. 

We use two variants of retrieve-then-generate fine-tuning, both using the special tokens \texttt{<RETRIEVAL>} and \texttt{</RETRIEVAL>} to delineate the retrieval step. In the first, \textbf{R}etrieve-\textbf{T}hen-\textbf{A}nswer (\textbf{RTA}), the model copies relevant passages. In the second, \textbf{C}ite-\textbf{C}ontext-\textbf{I}D (\textbf{CCI}), the model generates only the IDs of the relevant passages. Figure~\ref{fig:prompt-template} shows the templates for all variations.

\subsection{Retrieval Attention Probing}\label{sec:ret-attn-probe}

Our second proposal, \textbf{R}etrieval \textbf{A}ttention \textbf{P}robing (\textbf{RAP}), is an inference-time approach compatible with LCLMs without requiring re-training. Building on \citeauthor{wu2024retrievalHead}'s findings that specific attention heads are highly active during retrieval tasks (e.g., NIAH), RAP utilizes these retrieval-focused attention heads for context filtering before generating responses. For each attention head $h$ and query $q$, we track the Top-$M$ attention scores $A_h^M(q)$. $C_h^M(q)$, representing the $M$ passages corresponding to $A_h^M(q)$, are then selected, and the \textit{hit rate} for head $h$ is calculated as follows:
\begin{align}
    \text{HitRate}_h &= \frac{1}{N} \sum_{i=1}^{N} \frac{| C_h^M(q) \cap Z^*(q) |}{|Z^*(q)|},
\end{align}
where $N$ is the number of validation samples, and $Z^*(q)$ is the set of all relevant contexts given query $q$. Finally we select $Q$ heads with the top hit rates to be the retrieval heads as follows,
\begin{equation}
    \mathcal{H}_{\text{ret}} = \underset{\mathcal{H} \subseteq \{1, \dots, H\}, |\mathcal{H}_{\text{ret}}| = Q}{\arg\max} \sum_{h \in \mathcal{H}} \text{HitRate}_h,
\end{equation}
where $H$ is the number of attention heads, and $\mathcal{H}_{\text{ret}}$ is the set of \( Q \) attention heads with the top hit rates selected as the retrieval heads for context filtering.

Note that our definition of retrieval heads differs from \citep{wu2024retrievalHead} in two key ways: (1) Instead of focusing solely on the Top-$1$ passages, we allow each head to retain the Top-\( M \), which is particularly important for multi-hop reasoning tasks requiring multiple passages for reasoning. (2) We evaluate attention heads using the retrieval hit rate, a more direct metric for our downstream tasks.

During inference, we union all passages selected by all retrieval heads, $\mathcal{H}_{\text{ret}}$,
\begin{equation}
    C^* = \bigcup_{h \in \mathcal{H}_{\text{ret}}} \underset{c \in C}{\text{Top-}M} \, \alpha_h(c),
\end{equation}
\noindent where $\alpha_h(c)$ is the attention score of head $h$ on passage $c$, and $C^*$ is the set of the Top-$M$ selected passages from all heads $\mathcal{H}_{\text{ret}}$. We use $C^*$ to form a new CiC-style prompt and proceed to generating the final response. Since $|C^*| \ll |C|$, the final decoding is actually performed on a filtered contextual knowledge base with a much smaller length.

\subsection{Joint Retrieval Head Training}\label{sec:ret-head-modeling}

Our final proposal introduces a dedicated retrieval head to the LCLM model architecture. During inference, the model first uses the head to identify relevant passages, after which the generation head decodes a response conditioned on the retrieved content. During training, the retrieval and generation heads are jointly optimized using the Gumbel-TopK trick \citep{kool2019gumbel-topK}, which mitigates the non-differentiability of the retrieval process.

Figure~\ref{fig:retrieval-head-modeling} illustrates the modified model architecture. The retrieval head generates a binary mask, \( M \in \{0, 1\}^{|C|} \), indicating which passages to select (1) or ignore (0). The selected passages are then passed to the generation head for response generation. The retrieval head consists of: (1) encoders for the query and passage, using the LCLM's final hidden states, and (2) a scoring layer that computes relevance scores by concatenating their encoded vectors. The top \( K \) passages are then selected for the generation head to produce a response.

More specifically, let \( q \) denote a query and \( C = \{c_1, c_2, \dots, c_n\} \) be the set of \( n \) passages. For each pair \( (q, c_i) \), two single-layer encoders pool the final hidden states from an LCLM and generate a query vector, \( \textbf{h}_Q \), and a passage vector, \( \textbf{h}_{c_i} \):
\[
\textbf{h}_q = \text{enc}_q(q); \quad \textbf{h}_{c_i} = \text{enc}_c(c_i),
\]
Each pair \( (\textbf{h}_q, \textbf{h}_{c_i}) \) is then concatenated to form an input vector \( \textbf{v}_i = [\textbf{h}_q; \textbf{h}_{c_i}] \). A single scoring layer then predicts a scalar relevance score \( s_i \) for each pair, essentially computing $s_i = f(\textbf{v}_i)$ where \( f(\cdot) \) is our scoring function. With the set of scores \( S = \{\dots s_i \dots\} \), we can finally identify the indices $T$ of the selected passages as the ones that receive the top $K$ scores,
\[
T = \text{TopK}(S, K).
\]
We then set the binary mask $M$ for filtering the original context $C$ down to $C^*$, which is then used to prompt the LCLM to produce the final response.

\section{Experiment}
\label{sec:expt}

\textbf{Benchmarks.} 
We use LOFT \citep{lee2024loft} and our \icr{} benchmarks to evaluate LCLMs' in-context retrieval and reasoning capabilities. LOFT tests retrieval, single- and multi-hop question answering, and reasoning using NaturalQuestions \citep{kwiatkowski-etal-2019-natural}, HotpotQA \citep{yang2018hotpotqa}, and MuSiQue \citep{trivedi2022musique}. \icr{} uses NaturalQuestions and HotpotQA, and additionally includes FEVER \citep{thorne2018fever} for fact verification and WoW \citep{dinanwizard} for dialogue completion. Similar to LOFT, we report average scores across 100 test cases per task using the 32K context length versions, which is the maximum supported by all tested LCLMs.

\noindent \textbf{Metrics.}
For the question answering and fact verification tasks, we use the exact match in \citep{lee2024loft, adlakha2024evaluating}. We use ROUGE \citep{lin2004rouge} to assess on dialogue completion.

\noindent \textbf{Training Details.} 
We use \icr{}'s training set to fine-tune all models. Specifically, we randomly sample 7500, 7500, 5000, and 5000 instances for NaturalQuestions, HotpotQA, FEVER, and WoW, respectively. To verify the effectiveness of our proposed methods, we focus our experiments on Mistral-Instruct-7B model \citep{jiang2023mistral}. 

\noindent \textbf{Baselines.} 
We compare our approaches with the baselines in Sec.~\ref{sec:lclm-performance-icr}: Vanilla RAG, Closed-book, and Oracle RAG. 
We include the traditional ranking pipeline where we use TinyBERT \citep{jiao2020tinybert} and MiniLM \citep{wang2020minilm} fine-tuned on MS Marco passage retrieval corpus \citep{bajaj2016msmarco} to select the top-relevant passages for LCLMs. We report their retrieval performance in Appendix~\ref{appendix:reranker-retrieval-performance}.
We include the direct-answer SFT (SFT-DA) which follows \citeauthor{zhang2024raft} by concatenating the confounders with gold documents for SFT, and our methods in Sec.~\ref{sec:eliciting_icr} include two retrieve-then-generate SFT variants — Retrieve-then-Answer (SFT-RTA), and Cite-Context-ID (SFT-CCI) — and the joint retrieval head training (RetHead) and retrieval attention probing (RAP). 

\subsection{Main Results}
\label{sec:main-exp-result}

\begin{table*}[t]
\centering

\resizebox{0.8\linewidth}{!}{
\setlength{\tabcolsep}{6pt} %
\renewcommand{\arraystretch}{0.9} %
\begin{tabular}{lccccccc}
\toprule
\multicolumn{1}{c}{\textbf{Models \& Methods}} & \multicolumn{3}{c}{\textbf{LOFT}} & \multicolumn{4}{c}{\textbf{\icr}} \\
\cmidrule(lr){2-4} \cmidrule(lr){5-8}
& \textbf{NQ} & \textbf{HotpotQA} & \textbf{MUSIQUE} & \textbf{NQ} & \textbf{HotpotQA} & \textbf{FEVER} & \textbf{WoW} \\
\midrule
\textbf{GPT-4} w/ &  &  &  &  &  &  &  \\
{Close-book} & 0.58 & 0.57 & 0.28 & 0.61 & 0.50 & 0.85 & 0.16 \\
\cellcolor{gray!20}{Vanilla RAG} & \cellcolor{gray!20}0.85 & \cellcolor{gray!20}0.78 & \cellcolor{gray!20}0.51 & \cellcolor{gray!20}0.67 & \cellcolor{gray!20}0.64 & \cellcolor{gray!20}0.92 & \cellcolor{gray!20}0.15 \\
{Oracle} & 0.88 & 0.87 & 0.72 & 0.79 & 0.82 & 0.94 & 0.18 \\ \midrule
\textbf{Mistral-7B} w/ &  &  &  &  &  &  &  \\
{Close-book} & 0.46 & 0.38 & 0.05 & 0.49 & 0.38 & 0.73 & 0.14 \\
{Vanilla RAG} & 0.64 & 0.62 & 0.27 & 0.53 & 0.50 & 0.81 & 0.13 \\
RAG w/ RTA Prompting     & 0.60 & 0.70 & 0.27 & 0.54 & 0.51 & 0.83 & 0.15 \\
\cdashline{1-8} 
\textit{Re-ranking Strategy} &  &  &  &  &  &  &  \\
w/ TinyBERT $(k=8)$	& 0.88	&0.78	&0.29	&0.52	&0.47	&0.88	&0.12 \\
w/ MiniLM $(k=8)$	&0.83	&0.77	&0.27	&0.51&	0.46	&0.86&	0.13 \\
w/ TinyBERT $(k=32)$	&\textbf{0.87}	&\textbf{0.87}	&0.33	&0.62	&0.51&	0.91&	0.13 \\
w/ MiniLM $(k=32)$	&0.84	&0.84&	0.39	&0.61&	0.47	&\textbf{0.92}&	0.12 \\

\cellcolor{gray!20}{Oracle RAG} & \cellcolor{gray!20}0.89 & \cellcolor{gray!20}0.81 & \cellcolor{gray!20}0.41 & \cellcolor{gray!20}0.83 & \cellcolor{gray!20}0.81 & \cellcolor{gray!20}0.94 & \cellcolor{gray!20}0.18 \\ \cdashline{1-8}
\textit{Supervised Fine-tuning} &  &  &  &  &  &  &  \\
{SFT-Direct Answer} & \textcolor{myblue}{0.70} & \textcolor{myblue}{0.65} & \textcolor{myred}{0.25} & \textcolor{myblue}{0.59} & \textcolor{myblue}{0.70} & \textcolor{myblue}{0.90} & \textcolor{myblue}{0.22} \\
{SFT-Retrieve-then-Answer} & \textcolor{myblue}{0.74} & \textcolor{myblue}{0.69} & \textcolor{myblue}{0.33} & \textcolor{myblue}{0.60} & \textcolor{myblue}{0.67} & \textcolor{myblue}{0.91} & \textcolor{myblue}{0.22} \\
{SFT-Cite-Context-ID} & \textcolor{myblue}{0.76} & \textcolor{myred}{0.54} & \textcolor{myblue}{0.35} & \textcolor{myblue}{0.63} & \textcolor{myblue}{0.63} & \textcolor{myblue}{0.89} & \textcolor{myblue}{0.21} \\
\cdashline{1-8} 
\textit{Joint Retrieval Head Training} &  &  &  &  &  &  &  \\
RetHead $w/ \mathcal{L}_{gen} + \mathcal{L}_{ret}$                    & \textcolor{myred}{0.15}        & \textcolor{myred}{0.13}          & \textcolor{myred}{0.07}          & \textcolor{myred}{0.48}	          & \textcolor{myblue}{0.54}	        & \textcolor{myblue}{0.9}	          & \textcolor{myblue}{0.21} \\
RetHead $w/ \mathcal{L}_{gen}$                   & -        & -          & -                 & \textcolor{myred}{0.28}        & \textcolor{myred}{0.25}          & \textcolor{myblue}{0.82}          & \textcolor{myblue}{0.18}                                      \\
RetHead $w/ \mathcal{L}_{ret}$                     & -        & -          & -                 & \textcolor{myred}{0.39}        & \textcolor{myred}{0.44}          & \textcolor{myblue}{0.82}          & \textcolor{myblue}{0.13}                                    \\
\cdashline{1-8} 
\textit{Retrieval-Attention Probing} &  &  &  &  &  &  &  \\
{SFT-DA w/ RAP} & \textcolor{myblue}{0.78} & \textcolor{myblue}{0.76} & \textcolor{myblue}{\textbf{0.47}} & \textcolor{myblue}{\textbf{0.64}} & \textcolor{myblue}{0.67} & \textcolor{myblue}{0.89} & \textcolor{myblue}{0.21} \\
{SFT-RTA w/ RAP} & \textcolor{myblue}{{0.85}} & \textcolor{myblue}{{0.79}} & \textcolor{myblue}{0.39} & \textcolor{myblue}{0.63} & \textcolor{myblue}{\textbf{0.71}} & \textcolor{myblue}{\textbf{0.92}} & \textcolor{myblue}{\textbf{0.23}} \\
\bottomrule
\end{tabular}%
}
\caption{Main results on LOFT \citep{lee2024loft} and \icr{} for our methods applied on Mistral-2-7B-Instruct \citep{jiang2023mistral}. We also report the GPT-4 performance in the top panel. We highlight the \textcolor{myblue}{improved} and \textcolor{myred}{worsen} performances compared with Vanilla RAG, and \textbf{bold} the best method based on Mistral-7B, except the Oracle.}
\label{tab:loft-icr2-main-result}
\vspace{-15pt}
\end{table*}

As shown in Table~\ref{tab:loft-icr2-main-result}, all SFT variants outperform the Vanilla RAG on both benchmarks, indicating that LCLMs struggle to effectively leverage context as a knowledge base for RAG tasks. Furthermore, the gap between SFT-DA and the Oracle RAG highlights that supervised fine-tuning alone is insufficient to achieve optimal results.

We first apply RTA-style generation as a prompt-only method (RTA Prompting). The resulting performance gains, though modest, demonstrate the potential of decoupling the generation process into separate retrieval and generation stages.
Among the SFT variants, SFT-RTA in general outperforms the others with an average improvement of 2\%. Specifically on LOFT benchmark, both SFT-RTA and SFT-CCI outperform SFT-DA with 6\% and 2\% improvement on average, respectively, indicating the retrieve-then-generate strategy helps.  On \icr{}, however, all SFT variants perform the same. This demonstrates that \icr{} is a more discriminative benchmark than LOFT. 

We apply RAP to all SFT models. 
On average, RAP enhances the models significantly, with SFT-DA + RAP improving by 6\% and SFT-RTA + RAP by 8\%. The best-performing approach, SFT-RTA + RAP, achieves notable gains on the challenging \icr{} benchmark, with improvements of 3\%, 4\%, 1\%, and 1\% on NaturalQuestions, HotpotQA, FEVER, and WoW, respectively. It also achieves top performance on 5 out of the 7 tasks, demonstrating its superiority.Remarkably, it achieves comparable performance with the state-of-the-art GPT-4-Turbo on LOFT and \icr{} while using a much smaller model. 
Finally, RAP decoding is more effective for SFT models than with the original model, as SFT better activates retrieval-specific attention heads for the approach (see Sec.~\ref{sec:attn-head-analysis}). To verify the generalization of methods across various LLMs, we conduct experiments with LLaMA-3-Instruct \citep{dubey2024llama-3-technical-report} in Appendix~\ref{appendix:llama-3-generalization}: we observe the similar improvements for retrieve-then-generate training, and RAP decoding.

Compared to the traditional pipeline, LCLMs can effectively unify in-context retrieval and generation, particularly on challenging benchmarks such as \icr. This highlights the advantage of an end-to-end approach, wherein LCLMs contextually select retrieved items based on both the query and the evolving generation.

For the joint retrieval head training (RetHead), we performed experiments on training only the generation head ($w/ \mathcal{L}_{gen}$), only the retreieval head ($w/ \mathcal{L}_{ret}$), and both ($w/ \mathcal{L}_{gen} + \mathcal{L}_{ret}$). The last variant achieves the best performance, outperforming the Vanilla RAG baseline on \icr{}. However, it is not comparable with the SFT variants.

\subsection{In-Context Retrieval Performance}
\label{sec:main-exp-retrieval-result}

\begin{table}[t]
\centering
\setlength{\tabcolsep}{4pt} %
\renewcommand{\arraystretch}{0.95} %
\resizebox{\linewidth}{!}{%
\begin{tabular}{lcccccccccc}
\toprule
 \multirow{2}{*}{\textbf{Methods}} & \multicolumn{4}{c}{\textbf{LOFT}} & \multicolumn{5}{c}{\textbf{\icr}}  \\
\cmidrule(lr){2-5} \cmidrule(lr){6-9}
 & \textbf{NQ} & \textbf{HPQA} & \textbf{MUSI.}  & \textbf{NQ} & \textbf{HPQA} & \textbf{FEV.} & \textbf{WoW}   \\ 
\midrule

\textbf{SFT-RTA}  & 0.81 & 0.83 & 0.37  & 0.58 & 0.78 & 0.64 & 0.43  \\
\textbf{SFT-CCI}  & 0.65 & 0.61 & 0.26  & 0.63 & 0.73 & \textbf{0.69} & 0.49   \\
\cdashlinelr{1-9}
\textbf{RetHead} & & & & & & & \\
\textbf{$w/ \mathcal{L}_{gen} + \mathcal{L}_{ret}$} & 0.01 & 0.01 & 0.01  & 0.36 & 0.44 & 0.55 & 0.51   \\
\textbf{$w/ \mathcal{L}_{gen}$} & - & - & -  & 0.00 & 0.00 & 0.00 & 0.02 \\
\textbf{$w/ \mathcal{L}_{ret}$} & - & - & -  & 0.33 & 0.49 & 0.51 & 0.49   \\
\cdashlinelr{1-9}
\textbf{RAP Decoding} & & & & & & & \\
\textbf{Vanilla RAG}  & 0.19 & 0.25 & 0.10  & 0.36 & 0.13 & 0.00 & 0.00   \\
\textbf{SFT-DA}  & \textbf{0.95} & 0.70 & 0.27  & \textbf{0.75} & 0.67 & 0.60 & \textbf{0.69} & \\
\textbf{SFT-RTA} & \textbf{0.95} & \textbf{0.96} & \textbf{0.39} & 0.61 & \textbf{0.85} & 0.60 & 0.41 & \\
\bottomrule
\end{tabular}%
}
\caption{Retrieval performance measured by recall rate for various methods using Mistral-2-7B.}
\label{tab:loft-icr2-retrieval-performance}
\vspace{-15pt}
\end{table}

Table~\ref{tab:loft-icr2-retrieval-performance} reports the \textit{recall rates} for all methods, meaning how likely the models retrieve the relevant provenances. For the SFT variants, we evaluate the retrieval predictions produced in the retrieval phase. For the joint retrieval head training approach (RetHead), we analyze the predictions from the retrieval head. For RAP decoding, we assess the passages identified by the selected attention heads.

We find a strong correlation between a model's recall rate and its downstream task performance, highlighting the importance of in-context retrieval ability for LCLMs. Our best approach, SFT-RTA + RAP, achieves the highest recall. 
In contrast, Vanilla RAG + RAP exhibits poor retrieval performance, consistent with its limited improvement on the downstream tasks. 
The near-random retrieval performance of joint training of retrieval head in LOFT explains its failure of generalization.

\section{Discussion}

\subsection{Scaling the Supervised Fine-tuning}

\begin{figure*}
    \centering
    \includegraphics[width=0.85\linewidth]{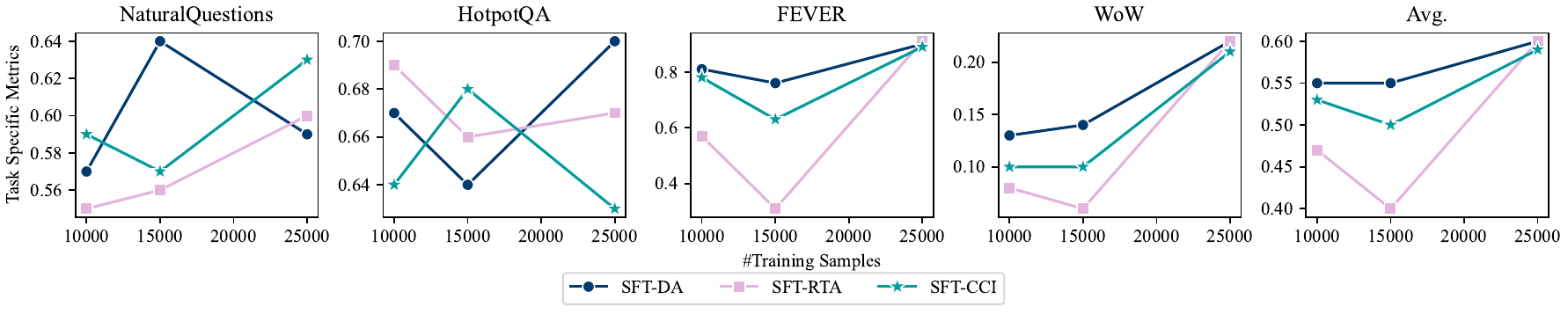}
    \vspace{-10pt}
    \caption{Scaling the number of training samples in \icr's training set.}
    \label{fig:scaling-data-size}
\vspace{-10pt}    
\end{figure*}

We are also interested in how the performance of the proposed SFT variants scale with the training set size. We train the three variants, SFT-DA, SFT-RTA, and SFT-CCI, with the same 10K, 15K and 25K examples from \icr{}'s training set, and report their performance on each task in the \icr{} benchmark, as shown in Fig~\ref{fig:scaling-data-size}. We observe that an increased training set size in general leads to an improved model performance. In particular, a smaller amount of training data fares worse with the retrieve-then-generate approaches, as they are by nature more challenging to learn compared to the SFT-DA approach, where answer is directly generated without an explicit retrieval step.

\subsection{Blocking Context Attention in Retrieve-then-generate Model}

\begin{table}[t]
\centering
\resizebox{0.7\linewidth}{!}{%
\begin{tabular}{lrrrr}
\toprule
\multirow{2}{*}{\textbf{\icr}} & \multicolumn{2}{c}{\textbf{SFT-RTA}}                   & \multicolumn{2}{c}{\textbf{Blocking Context}}        \\ \cmidrule{2-5} 
                     & \textbf{Metrics} & \multicolumn{1}{l}{\textbf{Recall}} & \textbf{Metrics} & \multicolumn{1}{l}{\textbf{Recall}} \\ \cmidrule{1-5} 
\textbf{NQ}          & 0.6              & 0.58                                & 0.61 \textcolor{myblue}{($\uparrow.01$)}            & 0.59 \textcolor{myblue}{($\uparrow.01$)}                               \\
\textbf{HPQA}    & 0.67             & 0.78                                & 0.68 \textcolor{myblue}{($\uparrow.01$)}            & 0.78 \textcolor{myblue}{($=.00$)}                               \\
\textbf{FEVER}       & 0.91             & 0.64                                & 0.91 \textcolor{myblue}{($=.00$)}            & 0.66 \textcolor{myblue}{($\uparrow.02$)}                               \\
\textbf{WoW}         & 0.22             & 0.43                                & 0.21 \textcolor{myred}{($\downarrow.01$)}            & 0.46 \textcolor{myblue}{($\uparrow.03$)}                               \\ \cdashlinelr{1-5} 
\textbf{Avg}         & 0.6              & 0.61          & 0.6 \textcolor{myblue}{($=.00$)}             & 0.62 \textcolor{myblue}{($\uparrow.01$)}         \\ \bottomrule
\end{tabular}}
\caption{SFT-RTA's performance before and after blocking the contextual knowledge base with attention mask. 
}
\label{tab:block-context-attention}
\vspace{-10pt}
\end{table}

To verify if models actually learn to generate the final responses only from the retrieval predictions in our retrieve-then-generate methods (Sec.~\ref{sec:retrieve-then-generate-tuning}), additional experiments were performed where we block attention to be paid onto the context beyond the retrieval predictions at the generation step.
Table~\ref{tab:block-context-attention} reports the impact of the blocking on the SFT-RTA variant. We observe the performance essentially stays the same, indicating the model indeed learned to generate largely on the retrieved passages, ignoring the original context.

\subsection{Effect on Attention Heads of Retrieve-then-Generate Fine-tuning}
\label{sec:attn-head-analysis}

\begin{figure}[!t]
    \centering
    \includegraphics[width=0.9\linewidth]{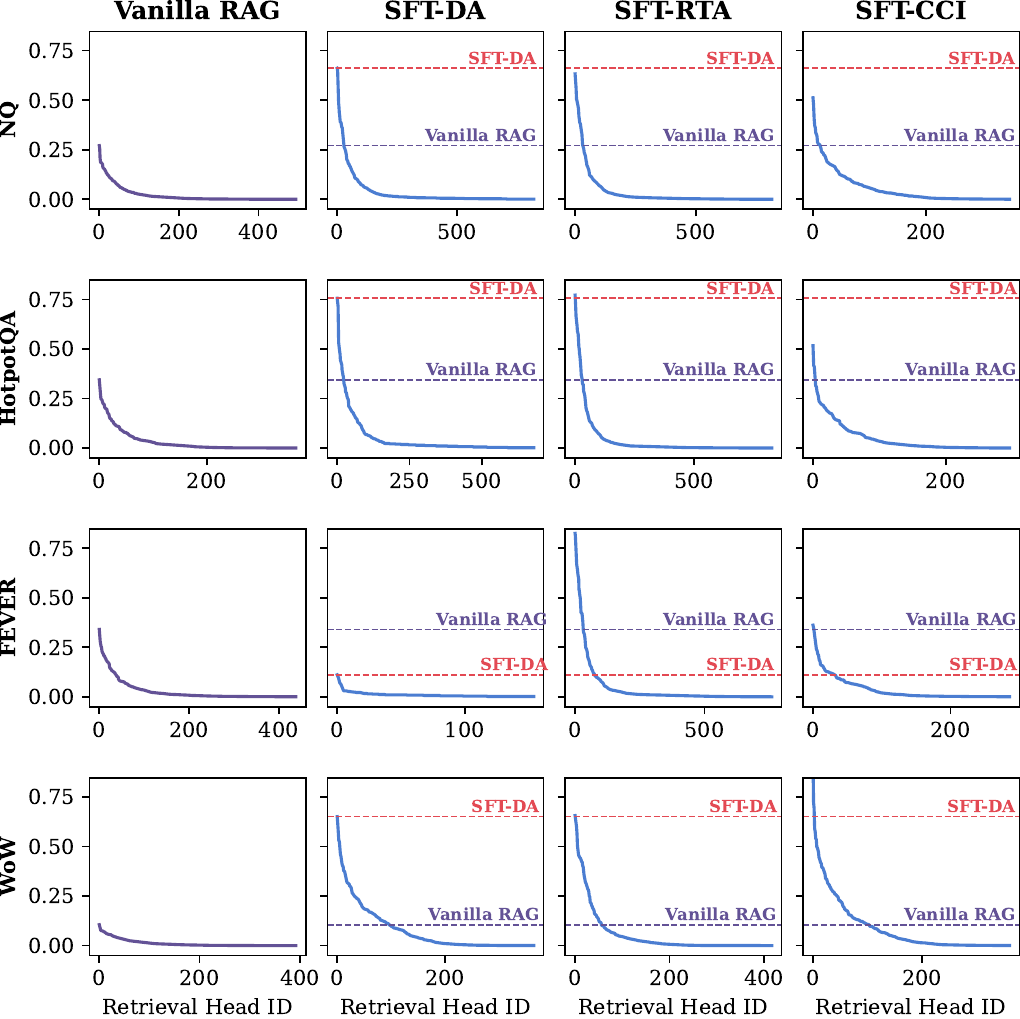}
    \caption{Attention heads with above-zero hit rates. 
    SFT produces more retrieval attention heads. Retrieve-then-generate training activates a higher peak of hit rate.
    }
    \label{fig:attention-head-score-analysis}
\vspace{-15pt}
\end{figure}

To understand if fine-tuning sharpens attention heads' focus on relevant passages, we compare the hit rates (Sec.~\ref{sec:ret-attn-probe}) achieved by the attention heads between Vanilla RAG and all of our SFT variants (Sec.~\ref{sec:retrieve-then-generate-tuning}), and the results are shown in Figure~\ref{fig:attention-head-score-analysis}. 

Similar to \citep{wu2024retrievalHead}, we find that a small group of attention heads can obtain higher hit rates than the others. However, unlike Vanilla RAG, SFT methods produce more retrieval-focused attention heads, and achieve higher peak hit rates. This demonstrates the effectiveness of our SFT methods and our curated \icr{} training set in enhancing LCLMs' performance of in-context retrieval. 

Among the three SFT variants, we find that SFT-RTA in general achieves higher peak hit rates and activates more attention heads for retrieval. In particular, SFT-CCI is not as effective in recruiting as many attention heads, possibly because context IDs themselves are not informative enough for the model to learn the retrieval task well. We also note that SFT-DA fares a lot worse on FEVER, possibly due to the lack of chain-of-thought style of assist. 

\subsection{Retrieval Attention Probing}

\begin{figure}[!t]
    \centering
    \begin{subfigure}[b]{0.9\linewidth}
        \centering
        \includegraphics[width=\linewidth]{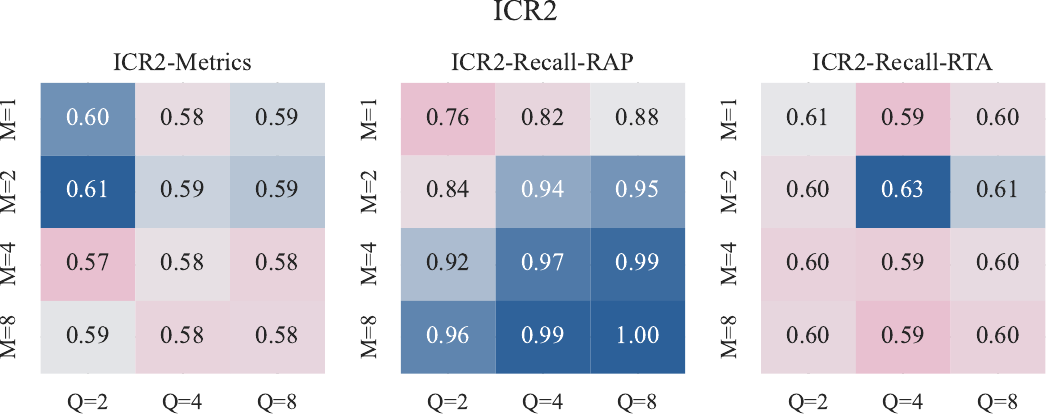}
        \label{fig:recall-attn}
    \end{subfigure}
    
    \vspace{-0.5cm} %
    
    \begin{subfigure}[b]{0.9\linewidth}
        \centering
        \includegraphics[width=\linewidth]{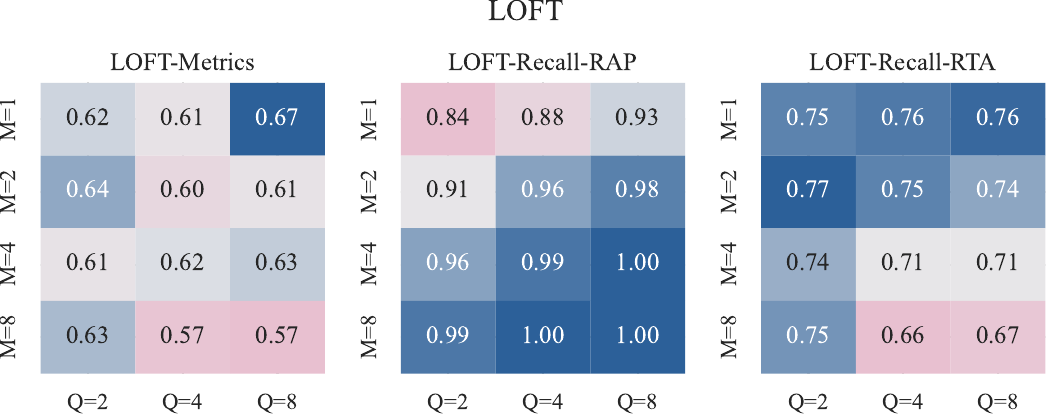}
        \label{fig:recall-rta}
    \end{subfigure}
    \vspace{-0.5cm}
    \caption{
    Effect of adjusting $Q$ (number of attention heads for retrieval) and $M$ (number of passages to select) when applying SFT-RTA + RAP. Left column: average model performance measured by the tasks-specific metrics. Middle column: attention heads' retrieval measured by the recall rate. Right column: RTA's retrieval measured by the exact match.
    }
    \label{fig:rap-perf-alter-Q-M}
\vspace{-10pt}
\end{figure}

Our inference-time method RAP (Sec.~\ref{sec:ret-attn-probe}) uses two hyperparameters: $Q$ is the number of the attention heads we recruit for retrieval, and $M$ is the number of passages each head retrieves. In this section, we apply different value settings when deploying the SFT-RTA + RAP combined approach to explore their effect on model performance. 

Figure~\ref{fig:rap-perf-alter-Q-M} shows the high-level results on both LOFT and \icr{} (Appendix~\ref{appendix:detailed-results-rap-decoding} has more details). As expected, we observe in the middle column that an increasing $M$ or $Q$ will increase the recall rate as the resulting larger pools of selected passages will more likely include the relevant ones. We also find that increasing $M$ as opposed to $Q$ is more effective in improving the recall rate as certain task such as HotpotQA requires multi-hop retrieval. However, the increased recall rate does not always translate to a better performance (left column) or RTA's recall rate
, as a higher $M$ or $Q$ may also introduce more confounders. This suggests the further reduction in confounding effects is still an opening future work.

\subsection{Decoding Speed}

\begin{table}[t]
\centering
\small
\begin{tabular}{lrr}
\toprule
\textbf{Benchmark}         & \cellcolor{gray!20} \textbf{SFT-DA} & \textbf{SFT-DA + RAP} \\ \midrule
NQ       & \cellcolor{gray!20} 4.23       & 4.99 \textcolor{myred}{($\uparrow.76$)}  \\
HotpotQA & \cellcolor{gray!20} 4.19       & 4.81 \textcolor{myred}{($\uparrow.62$)}  \\
MUSIQUE  & \cellcolor{gray!20} 4.24       & 5.05  \textcolor{myred}{($\uparrow.81$)} \\
\bottomrule
\end{tabular}
\caption{Latency for SFT-DA with and without the RAP decoding. Time is measured in seconds per query.}
\label{tab:rap-decoding-speed}
\vspace{-15pt}
\end{table}

Our final analysis is on the efficiency of RAP. Based on the SFT-DA variant, we report its latency with and without the RAP enhancement in Table~\ref{tab:rap-decoding-speed}. We find that the RAP decoding does not increase latency significantly, despite it adds one additional decoding step to the base method. This can be attributed to the much shorter context retrieved by the attention heads, thus avoiding expensive long-context computation as in the baseline. This increase can be further reduced with approaches such as KV caching, which we leave for future work.

\section{Conclusion}

In this paper, we introduce \icr{}, a new benchmark designed as a more discriminative benchmark for evaluating LCLMs in in-context retrieval and reasoning. 
Our findings highlight the limitations for the current models.
We propose three methods —retrieve-then-generate fine-tuning, retrieval attention probing, and joint retrieval head training — to enhance models, achieving the results comparable to GPT-4 with a smaller model footprint.

\section*{Ethical Considerations}

We do not expect any ethical concerns to be raised with respect to this work.

\section*{Limitations}

We acknowledge several limitations in this work. First, most experiments were conducted with a context length of 32K tokens. Our findings indicate that while many LCLMs claim to support longer contexts, their performance on tasks with 32K tokens remains suboptimal. Future work could focus on extending \icr{} and the proposed approaches to effectively support scenarios with longer context lengths.

Second, while Joint Retrieval Head Training demonstrates improved performance compared to Vanilla RAG, it still falls short of the performance achieved by the SFT variant. Future research could explore improved architectural designs to better integrate the supervision signals from both the retrieval and generation tasks.

Finally, our evaluation primarily utilizes the Mistral-7B model due to computational constraints. Extending the proposed methods to other LCLMs would provide a broader assessment of their generalization capabilities and effectiveness across different model architectures.

\bibliography{custom}

\begin{thebibliography}{52}
\providecommand{\natexlab}[1]{#1}

\bibitem[{Abdin et~al.(2024)Abdin, Jacobs, Awan, Aneja, Awadallah, Awadalla, Bach, Bahree, Bakhtiari, Behl et~al.}]{abdin2024phi}
Marah Abdin, Sam~Ade Jacobs, Ammar~Ahmad Awan, Jyoti Aneja, Ahmed Awadallah, Hany Awadalla, Nguyen Bach, Amit Bahree, Arash Bakhtiari, Harkirat Behl, et~al. 2024.
\newblock Phi-3 technical report: A highly capable language model locally on your phone.
\newblock \emph{arXiv preprint arXiv:2404.14219}.

\bibitem[{Achiam et~al.(2023)Achiam, Adler, Agarwal, Ahmad, Akkaya, Aleman, Almeida, Altenschmidt, Altman, Anadkat et~al.}]{achiam2023gpt}
Josh Achiam, Steven Adler, Sandhini Agarwal, Lama Ahmad, Ilge Akkaya, Florencia~Leoni Aleman, Diogo Almeida, Janko Altenschmidt, Sam Altman, Shyamal Anadkat, et~al. 2023.
\newblock Gpt-4 technical report.
\newblock \emph{arXiv preprint arXiv:2303.08774}.

\bibitem[{Adlakha et~al.(2024)Adlakha, BehnamGhader, Lu, Meade, and Reddy}]{adlakha2024evaluating}
Vaibhav Adlakha, Parishad BehnamGhader, Xing~Han Lu, Nicholas Meade, and Siva Reddy. 2024.
\newblock Evaluating correctness and faithfulness of instruction-following models for question answering.
\newblock \emph{Transactions of the Association for Computational Linguistics}, 12:775--793.

\bibitem[{Bai et~al.(2024)Bai, Lv, Zhang, Lyu, Tang, Huang, Du, Liu, Zeng, Hou, Dong, Tang, and Li}]{bai-etal-2024-longbench}
Yushi Bai, Xin Lv, Jiajie Zhang, Hongchang Lyu, Jiankai Tang, Zhidian Huang, Zhengxiao Du, Xiao Liu, Aohan Zeng, Lei Hou, Yuxiao Dong, Jie Tang, and Juanzi Li. 2024.
\newblock \href {https://doi.org/10.18653/v1/2024.acl-long.172} {{L}ong{B}ench: A bilingual, multitask benchmark for long context understanding}.
\newblock In \emph{Proceedings of the 62nd Annual Meeting of the Association for Computational Linguistics (Volume 1: Long Papers)}, pages 3119--3137, Bangkok, Thailand. Association for Computational Linguistics.

\bibitem[{Bajaj et~al.(2016)Bajaj, Campos, Craswell, Deng, Gao, Liu, Majumder, McNamara, Mitra, Nguyen et~al.}]{bajaj2016msmarco}
Payal Bajaj, Daniel Campos, Nick Craswell, Li~Deng, Jianfeng Gao, Xiaodong Liu, Rangan Majumder, Andrew McNamara, Bhaskar Mitra, Tri Nguyen, et~al. 2016.
\newblock Ms marco: A human generated machine reading comprehension dataset.
\newblock \emph{arXiv preprint arXiv:1611.09268}.

\bibitem[{BehnamGhader et~al.(2024)BehnamGhader, Adlakha, Mosbach, Bahdanau, Chapados, and Reddy}]{behnamghader2024llm2vec}
Parishad BehnamGhader, Vaibhav Adlakha, Marius Mosbach, Dzmitry Bahdanau, Nicolas Chapados, and Siva Reddy. 2024.
\newblock Llm2vec: Large language models are secretly powerful text encoders.
\newblock \emph{arXiv preprint arXiv:2404.05961}.

\bibitem[{Beltagy et~al.(2020)Beltagy, Peters, and Cohan}]{Beltagy2020Longformer}
Iz~Beltagy, Matthew~E. Peters, and Arman Cohan. 2020.
\newblock Longformer: The long-document transformer.
\newblock \emph{arXiv:2004.05150}.

\bibitem[{Chen et~al.(2017)Chen, Fisch, Weston, and Bordes}]{chen2017reading-drqa}
Danqi Chen, Adam Fisch, Jason Weston, and Antoine Bordes. 2017.
\newblock Reading {Wikipedia} to answer open-domain questions.
\newblock In \emph{Association for Computational Linguistics (ACL)}.

\bibitem[{Chen et~al.(2024)Chen, Qian, Tang, Lai, Liu, Han, and Jia}]{chenlonglora}
Yukang Chen, Shengju Qian, Haotian Tang, Xin Lai, Zhijian Liu, Song Han, and Jiaya Jia. 2024.
\newblock Longlora: Efficient fine-tuning of long-context large language models.
\newblock In \emph{The Twelfth International Conference on Learning Representations}.

\bibitem[{Dinan et~al.(2018)Dinan, Roller, Shuster, Fan, Auli, and Weston}]{dinanwizard}
Emily Dinan, Stephen Roller, Kurt Shuster, Angela Fan, Michael Auli, and Jason Weston. 2018.
\newblock Wizard of wikipedia: Knowledge-powered conversational agents.
\newblock In \emph{International Conference on Learning Representations}.

\bibitem[{Ding et~al.(2024)Ding, Zhang, Zhang, Xu, Shang, Xu, Yang, and Yang}]{dinglongrope}
Yiran Ding, Li~Lyna Zhang, Chengruidong Zhang, Yuanyuan Xu, Ning Shang, Jiahang Xu, Fan Yang, and Mao Yang. 2024.
\newblock Longrope: Extending llm context window beyond 2 million tokens.
\newblock In \emph{Forty-first International Conference on Machine Learning}.

\bibitem[{Dubey et~al.(2024)Dubey, Jauhri, Pandey, Kadian, Al-Dahle, Letman, Mathur, Schelten, Yang, Fan et~al.}]{dubey2024llama-3-technical-report}
Abhimanyu Dubey, Abhinav Jauhri, Abhinav Pandey, Abhishek Kadian, Ahmad Al-Dahle, Aiesha Letman, Akhil Mathur, Alan Schelten, Amy Yang, Angela Fan, et~al. 2024.
\newblock The llama 3 herd of models.
\newblock \emph{arXiv preprint arXiv:2407.21783}.

\bibitem[{Fu et~al.(2024)Fu, Panda, Niu, Yue, Hajishirzi, Kim, and Peng}]{fu-data-engineering-llm}
Yao Fu, Rameswar Panda, Xinyao Niu, Xiang Yue, Hannaneh Hajishirzi, Yoon Kim, and Hao Peng. 2024.
\newblock Data engineering for scaling language models to 128k context.
\newblock In \emph{Forty-first International Conference on Machine Learning}.

\bibitem[{Hsieh et~al.(2023)Hsieh, Li, Yeh, Nakhost, Fujii, Ratner, Krishna, Lee, and Pfister}]{hsieh-etal-2023-distilling}
Cheng-Yu Hsieh, Chun-Liang Li, Chih-kuan Yeh, Hootan Nakhost, Yasuhisa Fujii, Alex Ratner, Ranjay Krishna, Chen-Yu Lee, and Tomas Pfister. 2023.
\newblock \href {https://doi.org/10.18653/v1/2023.findings-acl.507} {Distilling step-by-step! outperforming larger language models with less training data and smaller model sizes}.
\newblock In \emph{Findings of the Association for Computational Linguistics: ACL 2023}, pages 8003--8017, Toronto, Canada. Association for Computational Linguistics.

\bibitem[{Izacard et~al.()Izacard, Caron, Hosseini, Riedel, Bojanowski, Joulin, and Grave}]{izacard-contriever}
Gautier Izacard, Mathilde Caron, Lucas Hosseini, Sebastian Riedel, Piotr Bojanowski, Armand Joulin, and Edouard Grave.
\newblock Unsupervised dense information retrieval with contrastive learning.
\newblock \emph{Transactions on Machine Learning Research}.

\bibitem[{Jiang et~al.(2023)Jiang, Sablayrolles, Mensch, Bamford, Chaplot, Casas, Bressand, Lengyel, Lample, Saulnier et~al.}]{jiang2023mistral}
Albert~Q Jiang, Alexandre Sablayrolles, Arthur Mensch, Chris Bamford, Devendra~Singh Chaplot, Diego de~las Casas, Florian Bressand, Gianna Lengyel, Guillaume Lample, Lucile Saulnier, et~al. 2023.
\newblock Mistral 7b.
\newblock \emph{arXiv preprint arXiv:2310.06825}.

\bibitem[{Jiao et~al.(2020)Jiao, Yin, Shang, Jiang, Chen, Li, Wang, and Liu}]{jiao2020tinybert}
Xiaoqi Jiao, Yichun Yin, Lifeng Shang, Xin Jiang, Xiao Chen, Linlin Li, Fang Wang, and Qun Liu. 2020.
\newblock Tinybert: Distilling bert for natural language understanding.
\newblock In \emph{Findings of the Association for Computational Linguistics: EMNLP 2020}, pages 4163--4174.

\bibitem[{Jin et~al.(2024)Jin, Han, Yang, Jiang, Liu, Chang, Chen, and Hu}]{jin2024llm}
Hongye Jin, Xiaotian Han, Jingfeng Yang, Zhimeng Jiang, Zirui Liu, Chia-Yuan Chang, Huiyuan Chen, and Xia Hu. 2024.
\newblock Llm maybe longlm: Self-extend llm context window without tuning.
\newblock \emph{CoRR}.

\bibitem[{Kamradt(2023)}]{niah_github}
Greg Kamradt. 2023.
\newblock Needle in a haystack - pressure testing llms.
\newblock \url{https://github.com/gkamradt/LLMTest_NeedleInAHaystack}.

\bibitem[{Karpukhin et~al.(2020)Karpukhin, Oguz, Min, Lewis, Wu, Edunov, Chen, and Yih}]{karpukhin2020DPR}
Vladimir Karpukhin, Barlas Oguz, Sewon Min, Patrick Lewis, Ledell Wu, Sergey Edunov, Danqi Chen, and Wen-tau Yih. 2020.
\newblock Dense passage retrieval for open-domain question answering.
\newblock In \emph{Proceedings of the 2020 Conference on Empirical Methods in Natural Language Processing (EMNLP)}, pages 6769--6781.

\bibitem[{Kim et~al.(2024)Kim, Chang, Karpinska, Garimella, Manjunatha, Lo, Goyal, and Iyyer}]{kim2024fables}
Yekyung Kim, Yapei Chang, Marzena Karpinska, Aparna Garimella, Varun Manjunatha, Kyle Lo, Tanya Goyal, and Mohit Iyyer. 2024.
\newblock Fables: Evaluating faithfulness and content selection in book-length summarization.
\newblock \emph{arXiv preprint arXiv:2404.01261}.

\bibitem[{Kool et~al.(2019)Kool, Van~Hoof, and Welling}]{kool2019gumbel-topK}
Wouter Kool, Herke Van~Hoof, and Max Welling. 2019.
\newblock Stochastic beams and where to find them: The gumbel-top-k trick for sampling sequences without replacement.
\newblock In \emph{International Conference on Machine Learning}, pages 3499--3508. PMLR.

\bibitem[{Kwiatkowski et~al.(2019)Kwiatkowski, Palomaki, Redfield, Collins, Parikh, Alberti, Epstein, Polosukhin, Devlin, Lee, Toutanova, Jones, Kelcey, Chang, Dai, Uszkoreit, Le, and Petrov}]{kwiatkowski-etal-2019-natural}
Tom Kwiatkowski, Jennimaria Palomaki, Olivia Redfield, Michael Collins, Ankur Parikh, Chris Alberti, Danielle Epstein, Illia Polosukhin, Jacob Devlin, Kenton Lee, Kristina Toutanova, Llion Jones, Matthew Kelcey, Ming-Wei Chang, Andrew~M. Dai, Jakob Uszkoreit, Quoc Le, and Slav Petrov. 2019.
\newblock \href {https://doi.org/10.1162/tacl_a_00276} {Natural questions: A benchmark for question answering research}.
\newblock \emph{Transactions of the Association for Computational Linguistics}, 7:452--466.

\bibitem[{Lee et~al.(2024)Lee, Chen, Dai, Dua, Sachan, Boratko, Luan, Arnold, Perot, Dalmia et~al.}]{lee2024loft}
Jinhyuk Lee, Anthony Chen, Zhuyun Dai, Dheeru Dua, Devendra~Singh Sachan, Michael Boratko, Yi~Luan, S{\'e}bastien~MR Arnold, Vincent Perot, Siddharth Dalmia, et~al. 2024.
\newblock Can long-context language models subsume retrieval, rag, sql, and more?
\newblock \emph{arXiv preprint arXiv:2406.13121}.

\bibitem[{Li et~al.(2024)Li, Verga, Sen, Yang, Viswanathan, Lewis, Watanabe, and Su}]{li2024alr}
Huayang Li, Pat Verga, Priyanka Sen, Bowen Yang, Vijay Viswanathan, Patrick Lewis, Taro Watanabe, and Yixuan Su. 2024.
\newblock Alr: A retrieve-then-reason framework for long-context question answering.
\newblock \emph{arXiv preprint arXiv:2410.03227}.

\bibitem[{Lin(2004)}]{lin2004rouge}
Chin-Yew Lin. 2004.
\newblock Rouge: A package for automatic evaluation of summaries.
\newblock In \emph{Text summarization branches out}, pages 74--81.

\bibitem[{Lin et~al.(2023)Lin, Zala, Cho, and Bansal}]{lin2023videodirectorgpt}
Han Lin, Abhay Zala, Jaemin Cho, and Mohit Bansal. 2023.
\newblock Videodirectorgpt: Consistent multi-scene video generation via llm-guided planning.
\newblock \emph{arXiv preprint arXiv:2309.15091}.

\bibitem[{Liu et~al.(2024)Liu, Lin, Hewitt, Paranjape, Bevilacqua, Petroni, and Liang}]{liu-etal-2024-lost}
Nelson~F. Liu, Kevin Lin, John Hewitt, Ashwin Paranjape, Michele Bevilacqua, Fabio Petroni, and Percy Liang. 2024.
\newblock \href {https://doi.org/10.1162/tacl_a_00638} {Lost in the middle: How language models use long contexts}.
\newblock \emph{Transactions of the Association for Computational Linguistics}, 12:157--173.

\bibitem[{Ma et~al.(2024)Ma, Wang, Yang, Wei, and Lin}]{ma2024fine}
Xueguang Ma, Liang Wang, Nan Yang, Furu Wei, and Jimmy Lin. 2024.
\newblock Fine-tuning llama for multi-stage text retrieval.
\newblock In \emph{Proceedings of the 47th International ACM SIGIR Conference on Research and Development in Information Retrieval}, pages 2421--2425.

\bibitem[{Petroni et~al.(2021)Petroni, Piktus, Fan, Lewis, Yazdani, De~Cao, Thorne, Jernite, Karpukhin, Maillard, Plachouras, Rockt{\"a}schel, and Riedel}]{petroni-etal-2021-kilt}
Fabio Petroni, Aleksandra Piktus, Angela Fan, Patrick Lewis, Majid Yazdani, Nicola De~Cao, James Thorne, Yacine Jernite, Vladimir Karpukhin, Jean Maillard, Vassilis Plachouras, Tim Rockt{\"a}schel, and Sebastian Riedel. 2021.
\newblock \href {https://doi.org/10.18653/v1/2021.naacl-main.200} {{KILT}: a benchmark for knowledge intensive language tasks}.
\newblock In \emph{Proceedings of the 2021 Conference of the North American Chapter of the Association for Computational Linguistics: Human Language Technologies}, pages 2523--2544, Online. Association for Computational Linguistics.

\bibitem[{Qiu et~al.(2023)Qiu, Ziser, Korhonen, Ponti, and Cohen}]{qiu-etal-2023-detecting}
Yifu Qiu, Yftah Ziser, Anna Korhonen, Edoardo Ponti, and Shay Cohen. 2023.
\newblock \href {https://doi.org/10.18653/v1/2023.emnlp-main.551} {Detecting and mitigating hallucinations in multilingual summarisation}.
\newblock In \emph{Proceedings of the 2023 Conference on Empirical Methods in Natural Language Processing}, pages 8914--8932, Singapore. Association for Computational Linguistics.

\bibitem[{Robertson et~al.(2009)Robertson, Zaragoza et~al.}]{robertson2009probabilistic-bm25}
Stephen Robertson, Hugo Zaragoza, et~al. 2009.
\newblock The probabilistic relevance framework: Bm25 and beyond.
\newblock \emph{Foundations and Trends{\textregistered} in Information Retrieval}, 3(4):333--389.

\bibitem[{Robertson and Jones(1976)}]{robertson1976relevance-bm25}
Stephen~E Robertson and K~Sparck Jones. 1976.
\newblock Relevance weighting of search terms.
\newblock \emph{Journal of the American Society for Information science}, 27(3):129--146.

\bibitem[{Saxena and Keller(2024)}]{saxena-keller-2024-moviesum}
Rohit Saxena and Frank Keller. 2024.
\newblock \href {https://doi.org/10.18653/v1/2024.findings-acl.239} {{M}ovie{S}um: An abstractive summarization dataset for movie screenplays}.
\newblock In \emph{Findings of the Association for Computational Linguistics: ACL 2024}, pages 4043--4050, Bangkok, Thailand. Association for Computational Linguistics.

\bibitem[{Shridhar et~al.(2023)Shridhar, Stolfo, and Sachan}]{shridhar2023distilling}
Kumar Shridhar, Alessandro Stolfo, and Mrinmaya Sachan. 2023.
\newblock Distilling reasoning capabilities into smaller language models.
\newblock In \emph{Findings of the Association for Computational Linguistics: ACL 2023}, pages 7059--7073.

\bibitem[{Su et~al.(2024)Su, Ahmed, Lu, Pan, Bo, and Liu}]{su2024roformer}
Jianlin Su, Murtadha Ahmed, Yu~Lu, Shengfeng Pan, Wen Bo, and Yunfeng Liu. 2024.
\newblock Roformer: Enhanced transformer with rotary position embedding.
\newblock \emph{Neurocomputing}, 568:127063.

\bibitem[{Thorne et~al.(2018)Thorne, Vlachos, Christodoulopoulos, and Mittal}]{thorne2018fever}
James Thorne, Andreas Vlachos, Christos Christodoulopoulos, and Arpit Mittal. 2018.
\newblock Fever: a large-scale dataset for fact extraction and verification.
\newblock In \emph{Proceedings of the 2018 Conference of the North American Chapter of the Association for Computational Linguistics: Human Language Technologies, Volume 1 (Long Papers)}, pages 809--819.

\bibitem[{Trivedi et~al.(2022)Trivedi, Balasubramanian, Khot, and Sabharwal}]{trivedi2022musique}
Harsh Trivedi, Niranjan Balasubramanian, Tushar Khot, and Ashish Sabharwal. 2022.
\newblock Musique: Multihop questions via single-hop question composition.
\newblock \emph{Transactions of the Association for Computational Linguistics}, 10:539--554.

\bibitem[{Wang et~al.(2024{\natexlab{a}})Wang, Chen, Cheng, Liao, Zhang, Wu, Yu, Xu, Zhang, Luo, Li, Yang, Huang, and Li}]{wang-etal-2024-leave}
Minzheng Wang, Longze Chen, Fu~Cheng, Shengyi Liao, Xinghua Zhang, Bingli Wu, Haiyang Yu, Nan Xu, Lei Zhang, Run Luo, Yunshui Li, Min Yang, Fei Huang, and Yongbin Li. 2024{\natexlab{a}}.
\newblock \href {https://aclanthology.org/2024.emnlp-main.322} {Leave no document behind: Benchmarking long-context {LLM}s with extended multi-doc {QA}}.
\newblock In \emph{Proceedings of the 2024 Conference on Empirical Methods in Natural Language Processing}, pages 5627--5646, Miami, Florida, USA. Association for Computational Linguistics.

\bibitem[{Wang et~al.(2020)Wang, Wei, Dong, Bao, Yang, and Zhou}]{wang2020minilm}
Wenhui Wang, Furu Wei, Li~Dong, Hangbo Bao, Nan Yang, and Ming Zhou. 2020.
\newblock Minilm: Deep self-attention distillation for task-agnostic compression of pre-trained transformers.
\newblock \emph{Advances in neural information processing systems}, 33:5776--5788.

\bibitem[{Wang et~al.(2024{\natexlab{b}})Wang, Xiong, Zhou, Lin, Zhao, Kang, Feng, and Liu}]{wang2024loong}
Yuqing Wang, Tianwei Xiong, Daquan Zhou, Zhijie Lin, Yang Zhao, Bingyi Kang, Jiashi Feng, and Xihui Liu. 2024{\natexlab{b}}.
\newblock Loong: Generating minute-level long videos with autoregressive language models.
\newblock \emph{arXiv preprint arXiv:2410.02757}.

\bibitem[{Wu et~al.(2020)Wu, Petroni, Josifoski, Riedel, and Zettlemoyer}]{wu2019-BLINK}
Ledell Wu, Fabio Petroni, Martin Josifoski, Sebastian Riedel, and Luke Zettlemoyer. 2020.
\newblock Zero-shot entity linking with dense entity retrieval.
\newblock In \emph{EMNLP}.

\bibitem[{Wu et~al.(2024)Wu, Wang, Xiao, Peng, and Fu}]{wu2024retrievalHead}
Wenhao Wu, Yizhong Wang, Guangxuan Xiao, Hao Peng, and Yao Fu. 2024.
\newblock Retrieval head mechanistically explains long-context factuality.
\newblock \emph{arXiv preprint arXiv:2404.15574}.

\bibitem[{Xiong et~al.(2024)Xiong, Liu, Molybog, Zhang, Bhargava, Hou, Martin, Rungta, Sankararaman, Oguz et~al.}]{xiong2024effective}
Wenhan Xiong, Jingyu Liu, Igor Molybog, Hejia Zhang, Prajjwal Bhargava, Rui Hou, Louis Martin, Rashi Rungta, Karthik~Abinav Sankararaman, Barlas Oguz, et~al. 2024.
\newblock Effective long-context scaling of foundation models.
\newblock In \emph{Proceedings of the 2024 Conference of the North American Chapter of the Association for Computational Linguistics: Human Language Technologies (Volume 1: Long Papers)}, pages 4643--4663.

\bibitem[{Xue et~al.(2024)Xue, Chen, Li, Hu, Zhu, Li, Fang, Tang, Yang, Liu et~al.}]{xue2024longvila}
Fuzhao Xue, Yukang Chen, Dacheng Li, Qinghao Hu, Ligeng Zhu, Xiuyu Li, Yunhao Fang, Haotian Tang, Shang Yang, Zhijian Liu, et~al. 2024.
\newblock Longvila: Scaling long-context visual language models for long videos.
\newblock \emph{arXiv preprint arXiv:2408.10188}.

\bibitem[{Yang et~al.(2024)Yang, Yang, Hui, Zheng, Yu, Zhou, Li, Li, Liu, Huang et~al.}]{yang2024qwen2}
An~Yang, Baosong Yang, Binyuan Hui, Bo~Zheng, Bowen Yu, Chang Zhou, Chengpeng Li, Chengyuan Li, Dayiheng Liu, Fei Huang, et~al. 2024.
\newblock Qwen2 technical report.
\newblock \emph{arXiv preprint arXiv:2407.10671}.

\bibitem[{Yang et~al.(2018)Yang, Qi, Zhang, Bengio, Cohen, Salakhutdinov, and Manning}]{yang2018hotpotqa}
Zhilin Yang, Peng Qi, Saizheng Zhang, Yoshua Bengio, William Cohen, Ruslan Salakhutdinov, and Christopher~D Manning. 2018.
\newblock Hotpotqa: A dataset for diverse, explainable multi-hop question answering.
\newblock In \emph{Proceedings of the 2018 Conference on Empirical Methods in Natural Language Processing}, pages 2369--2380.

\bibitem[{Zaheer et~al.(2020)Zaheer, Guruganesh, Dubey, Ainslie, Alberti, Ontanon, Pham, Ravula, Wang, Yang et~al.}]{zaheer2020big}
Manzil Zaheer, Guru Guruganesh, Kumar~Avinava Dubey, Joshua Ainslie, Chris Alberti, Santiago Ontanon, Philip Pham, Anirudh Ravula, Qifan Wang, Li~Yang, et~al. 2020.
\newblock Big bird: Transformers for longer sequences.
\newblock \emph{Advances in neural information processing systems}, 33:17283--17297.

\bibitem[{Zhang et~al.(2023)Zhang, Lu, Islam, Wang, Yu, Bansal, and Bertasius}]{zhang2023simple}
Ce~Zhang, Taixi Lu, Md~Mohaiminul Islam, Ziyang Wang, Shoubin Yu, Mohit Bansal, and Gedas Bertasius. 2023.
\newblock A simple llm framework for long-range video question-answering.
\newblock \emph{arXiv preprint arXiv:2312.17235}.

\bibitem[{Zhang et~al.(2024{\natexlab{a}})Zhang, Patil, Jain, Shen, Zaharia, Stoica, and Gonzalez}]{zhang2024raft}
Tianjun Zhang, Shishir~G Patil, Naman Jain, Sheng Shen, Matei Zaharia, Ion Stoica, and Joseph~E Gonzalez. 2024{\natexlab{a}}.
\newblock Raft: Adapting language model to domain specific rag.
\newblock \emph{arXiv preprint arXiv:2403.10131}.

\bibitem[{Zhang et~al.(2024{\natexlab{b}})Zhang, Sun, Chen, Pfister, Zhang, and Arik}]{zhang2024chain}
Yusen Zhang, Ruoxi Sun, Yanfei Chen, Tomas Pfister, Rui Zhang, and Sercan~{\"O} Arik. 2024{\natexlab{b}}.
\newblock Chain of agents: Large language models collaborating on long-context tasks.
\newblock \emph{arXiv preprint arXiv:2406.02818}.

\bibitem[{Zhao et~al.(2024)Zhao, Zu, Xu, Lu, He, Ding, Gui, Zhang, and Huang}]{zhao2024longagent}
Jun Zhao, Can Zu, Hao Xu, Yi~Lu, Wei He, Yiwen Ding, Tao Gui, Qi~Zhang, and Xuanjing Huang. 2024.
\newblock Longagent: Scaling language models to 128k context through multi-agent collaboration.
\newblock \emph{arXiv preprint arXiv:2402.11550}.

\end{thebibliography}

\newpage

\appendix

\section{Statistics for LOFT and \icr{}}
\label{appendix:statistics-loft-icr}

\begin{table}[h]
\centering
\small
\begin{tabular}{lllll}
\toprule
                               & \textbf{Task}   & \textbf{\#CTX} & \textbf{\#Tokens} & \textbf{\#Prov} \\ \midrule
\multirow{3}{*}{\textbf{LOFT}} & \textbf{NQ}     & 215             & 28,911             & 1                        \\
                               & \textbf{HPQA}   & 276             & 28,924             & 1.98                        \\
                               & \textbf{MUSIQ.} & 207             & 28,806             & 1.67                        \\ \midrule
\multirow{4}{*}{\textbf{\icr{}}}  & \textbf{NQ}     & 202             & 20,441             & 1                        \\
                               & \textbf{HPQA}   & 202             & 20,818             & 2.09                        \\
                               & \textbf{FEVER}  & 202             & 21,136             & 1                        \\
                               & \textbf{WoW}    & 202             & 21,068             & 1                        \\ \bottomrule
\end{tabular}
\caption{Statistics for LOFT and our dataset \icr{}: \#CTX is to the average number of passages per query in the contextual knowledge base, \#Tokens is the average length of the CiC prompt, and \#Prov. is the average number of provenance (positive passages) per query.}
\label{tab:benchmark-statistics}
\vspace{-10pt}
\end{table}

We present the comparison in dataset statistics between LOFT and \icr{} in Table~\ref{tab:benchmark-statistics}.

\section{Prompt Template for the Retrieval-augmented Generation}
\label{appendix:rag-prompt-templates}

\begin{table*}[!th]
\centering
\resizebox{\linewidth}{!}{%
\begin{tabular}{cl}
\toprule
 \textbf{Task} & {\textbf{Prompt Template}} \\ \midrule
    \multirow{4}{*}{\rotatebox{90}{\textbf{QA}}} & {\fontfamily{lmtt}\selectfont [INST] Please answer the following question given the following passages:} \\ 
    & {\fontfamily{lmtt}\selectfont \{Corpus\}} \\
    & {\fontfamily{lmtt}\selectfont Question: \{Query\}} \\
    & {\fontfamily{lmtt}\selectfont Answer: [/INST]} \\ \midrule
    
    \multirow{5}{*}{\rotatebox{90}{\small \textbf{Fact Verification}}} 
 & {\fontfamily{lmtt}\selectfont [INST] According to the following passages, please verify the given claim} \\
   & {\fontfamily{lmtt}\selectfont and predict your judgment on its factuality as TRUE or FALSE:} \\ 
   & {\fontfamily{lmtt}\selectfont \{Corpus\}} \\
   & {\fontfamily{lmtt}\selectfont Claim: \{Query\}} \\
   & {\fontfamily{lmtt}\selectfont Judgement: [/INST]} \\ \midrule

    \multirow{7}{*}{\rotatebox{90}{\small \textbf{Dialogue Completion}}} 
    & {\fontfamily{lmtt}\selectfont [INST] According to the given passages, please provide a single response to} \\ 
    & {\fontfamily{lmtt}\selectfont complete the following conversation by role-playing as either Person A or} \\ 
    & {\fontfamily{lmtt}\selectfont Person B. Your response should be as knowledgeable and coherent with the} \\
    & {\fontfamily{lmtt}\selectfont conversation history as possible:} \\ 
    & {\fontfamily{lmtt}\selectfont \{Corpus\}} \\
    & {\fontfamily{lmtt}\selectfont Conversation: \{Query\}} \\
    & {\fontfamily{lmtt}\selectfont [/INST]} \\ \bottomrule
\end{tabular}}
\caption{Prompt template used for RAG tasks in our experiments. {\fontfamily{lmtt}\selectfont \{Corpus\}} refers to the provided contextual knowledge base, and {\fontfamily{lmtt}\selectfont \{Query\}} refers to a query in \icr{} or LOFT. A {\fontfamily{lmtt}\selectfont \{Query\}} can be a question in the question answering tasks \citep{kwiatkowski-etal-2019-natural,yang2018hotpotqa}, a claim to be verified in the fact verification task \citep{thorne2018fever}, or a conversation history in the dialogue completion task \citep{dinanwizard}.}
\label{tab:rag-prompt-templates}
\end{table*}

We show the Corpus-in-Context (CiC; \citealt{lee2024loft}) prompt template used in our experiments in Table~\ref{tab:rag-prompt-templates}.

\section{Generalization to Other LCLMs}
\label{appendix:llama-3-generalization}

\begin{table}[th]
\centering
\small
\begin{tabular}{lcccc}
\toprule
\multirow{2}{*}{\textbf{LLaMA-3-Instruct}} & \multicolumn{2}{c}{\textbf{LOFT}} & \multicolumn{2}{c}{\textbf{\icr}} \\
\cmidrule{2-5}
& \textbf{NQ} & \textbf{HPQA} & \textbf{NQ} & \textbf{HPQA} \\
\midrule
Closebook                & 0.41 & 0.19 & 0.44 & 0.19 \\
RAG                      & 0.45 & 0.56 & 0.53 & 0.45 \\
RAG w/ TinyBERT          & \textbf{0.89} & \textbf{0.89} & 0.55 & 0.48 \\
Oracle                   & 0.81 & 0.77 & 0.80 & 0.74 \\
\midrule
SFT-DA                   & 0.80 & 0.60 & 0.51 & 0.52 \\
SFT-RTA                  & 0.84 & 0.63 & 0.51 & \textbf{0.57} \\
SFT-DA w/ RAP            & 0.83 & 0.72 & \textbf{0.61} & 0.60 \\
\bottomrule
\end{tabular}
\caption{LLaMA-3's performances on NQ and HotpotQA for LOFT and \icr with our methods on NQ-LOFT and HPQA-LOFT.}
\label{tab:llama3-performance}
\end{table}

Table~\ref{tab:llama3-performance} presents the performance of the LLaMA-3-Instruct-8B model across two benchmarks—NQ-LOFT and HPQA-LOFT—evaluated under LOFT and \icr. The results highlight several key trends. First, the baseline Closebook performs poorly, particularly on the HPQA dataset, underscoring the need for external information. Incorporating retrieval via RAG offers significant improvements, especially on HPQA, where performance increases from 0.19 to 0.56. The Oracle setting represents the upper bound of retrieval quality, and shows the strongest performance overall.

We observe that adding a reranker such as TinyBERT to RAG further improves results under the LOFT setup, reaching 0.89 on both NQ and HPQA—approaching Oracle-level performance. Among supervised fine-tuning methods, SFT-DA and SFT-RTA provide moderate improvements over vanilla RAG. Notably, integrating our proposed RAP mechanism (e.g., in SFT-DA w/ RAP) leads to consistent gains, especially under \icr . For example, SFT-DA w/ RAP achieves the best NQ score (0.61), validating the effectiveness of RAP in enhancing decision quality during retrieval and answer generation.

\section{Retrieval Performance for Re-ranking Strategies}
\label{appendix:reranker-retrieval-performance}
\begin{table*}[t]
\centering
\small
\setlength{\tabcolsep}{6pt}
\renewcommand{\arraystretch}{1.2}
\begin{tabular}{l|c|ccc|cccc}
\toprule
\multicolumn{1}{c}{\multirow{2}{*}{\textbf{Re-ranker Model}}} & \multicolumn{1}{c}{\multirow{2}{*}{\textbf{K}}} & \multicolumn{3}{c}{\textbf{LOFT}} & \multicolumn{4}{c}{\textbf{\icr}} \\
\cmidrule(lr){3-5} \cmidrule(lr){6-9}
& & NQ & HPQA & MUSI. & NQ & HPQA & FEV & WoW \\

\midrule

\rowcolor{gray!5}
TinyBERT            & 8   & 0.99 & 0.98 & 0.80    & 0.61 & 0.65  & 0.68 & 0.35 \\
MiniLM              & 8   & \textbf{1.00} & 0.98 & 0.82    & 0.66 & 0.65  & 0.70 & 0.27 \\
\rowcolor{gray!5}
TinyBERT            & 32  & 0.99 & 0.99 & 0.81    & 0.89 & 0.82  & 0.87 & 0.55 \\
MiniLM              & 32  & \textbf{1.00} & \textbf{1.00} & 0.90    & 0.86 & 0.81  & 0.89 & 0.43 \\
\rowcolor{gray!5}
TinyBERT            & 50  & 0.99 & 0.99 & 0.82    & \textbf{0.91} & 0.84  & 0.92 & 0.60 \\
MiniLM              & 50  & \textbf{1.00} & \textbf{1.00} & 0.92    & \textbf{0.91} & 0.88  & 0.91 & 0.50 \\
\midrule
SFT-RTA w/ RAP                & /   & 0.95 & 0.99 & \textbf{1.00} & 0.90 & \textbf{0.95} & \textbf{0.97} & \textbf{0.92} \\
\bottomrule
\end{tabular}
\caption{Retrieval performance comparison of re-ranker models across LOFT and \icr benchmarks.}
\label{tab:reranker-retrieval-performance}
\end{table*}

Table~\ref{tab:reranker-retrieval-performance} presents a comprehensive evaluation of different re-ranking strategies applied to retrieval-augmented generation (RAG) across two benchmark suites: LOFT (comprising NQ, HPQA, and MUSI.) and \icr (including NQ, HPQA, FEV, and WoW). We compare baseline RAG without re-ranking, re-ranking using lightweight models (TinyBERT and MiniLM) at varying retrieval depths $k\in \{8,32,50\}$, and our best-performing variant (SFT-RTA w/ RAP). Results show consistent improvements with increased $k$, and MiniLM-based re-rankers often slightly outperform TinyBERT counterparts. Notably, the SFT-RTA w/ RAP model yields the highest performance across most datasets, particularly excelling in more challenging \icr domains. These findings suggest the strong retrieval capabilities of our SFT-RTA approach compared to the traditional re-ranking strategies.

\section{Detailed Results for RAP Decoding}
\label{appendix:detailed-results-rap-decoding}

\begin{figure*}[t!]
    \centering
    \begin{subfigure}{0.6\linewidth}
        \centering
        \includegraphics[width=\linewidth]{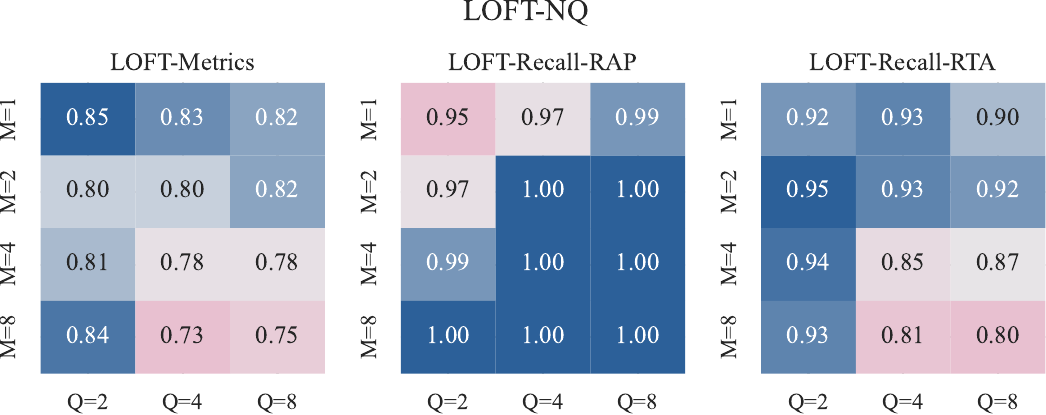}
        \label{fig:subfig1}
    \end{subfigure}
    
    \begin{subfigure}{0.6\linewidth}
        \centering
        \includegraphics[width=\linewidth]{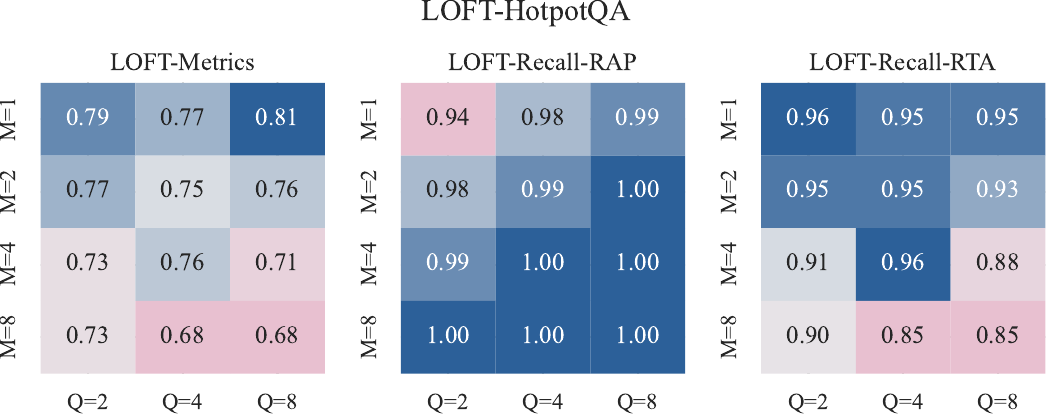}
        \label{fig:subfig2}
    \end{subfigure}
    
    \begin{subfigure}{0.6\linewidth}
        \centering
        \includegraphics[width=\linewidth]{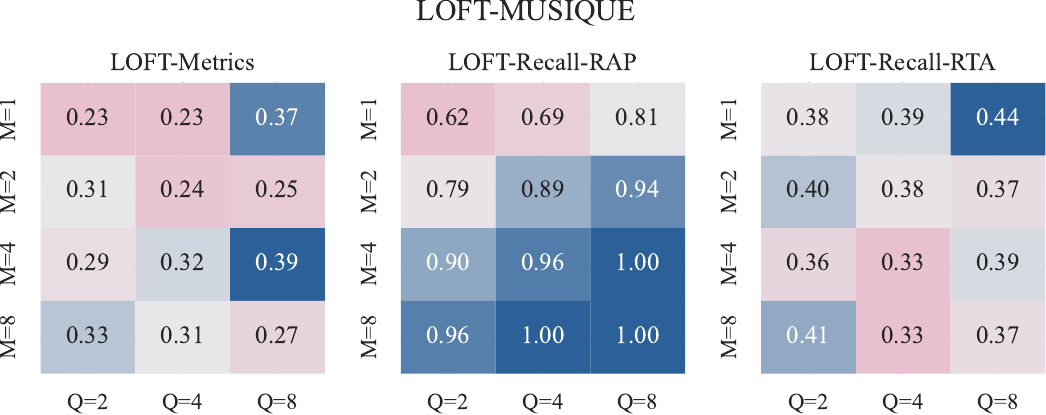}
        \label{fig:subfig3}
    \end{subfigure}
    
    \caption{
Effect of adjusting $Q$ (number of attention heads for retrieval) and $M$ (number of passages to select) when applying SFT-RTA + RAP on all tasks in LOFT. Left column: average model performance measured by the tasks-specific metrics. Middle column: retrieval performance measured by the recall rate. Right column: retrieval performance measured by the exact match.
    }
    \label{fig:detailed-loft-rap-results}
\end{figure*}

\begin{figure*}[t!]
    \centering
    \begin{subfigure}{0.6\linewidth}
        \centering
        \includegraphics[width=\linewidth]{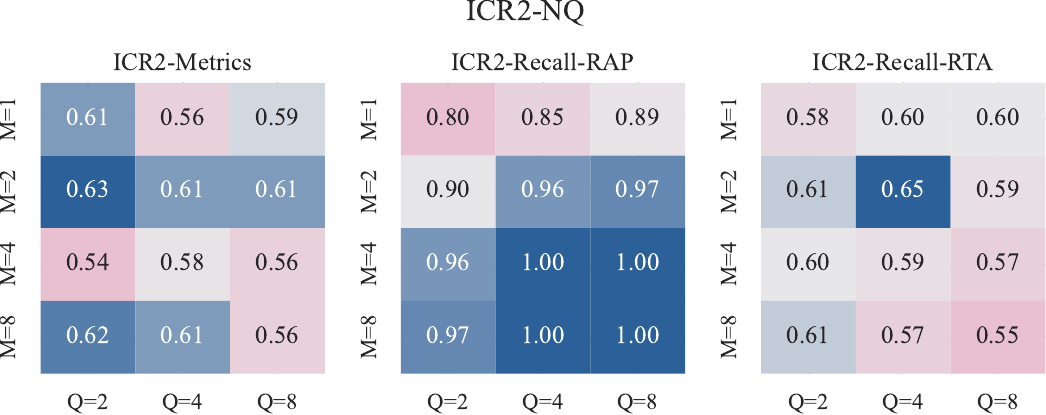}
        \label{fig:subfig4}
    \end{subfigure}
    
    \begin{subfigure}{0.6\linewidth}
        \centering
        \includegraphics[width=\linewidth]{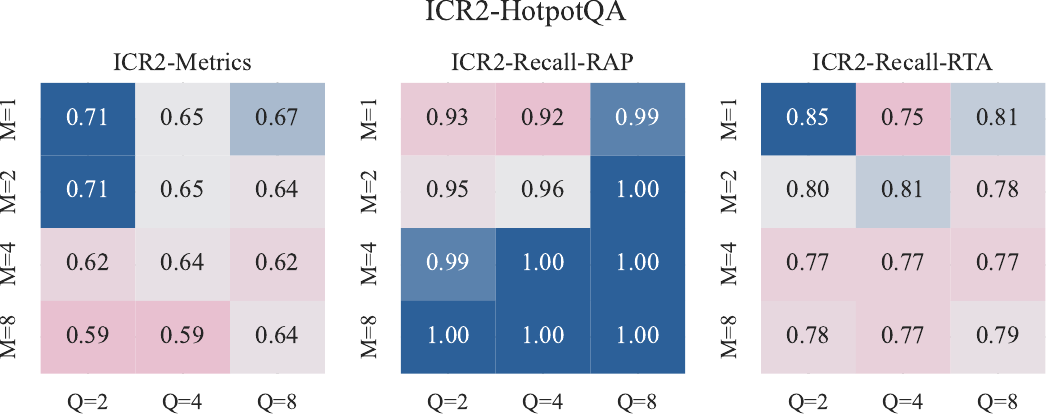}
        \label{fig:subfig5}
    \end{subfigure}
    
    \begin{subfigure}{0.6\linewidth}
        \centering
        \includegraphics[width=\linewidth]{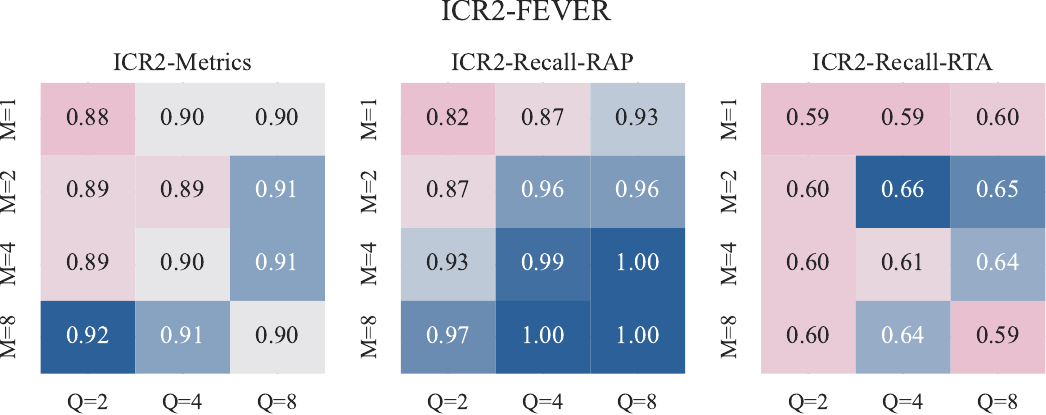}
        \label{fig:subfig6}
    \end{subfigure}
    
    \begin{subfigure}{0.6\linewidth}
        \centering
        \includegraphics[width=\linewidth]{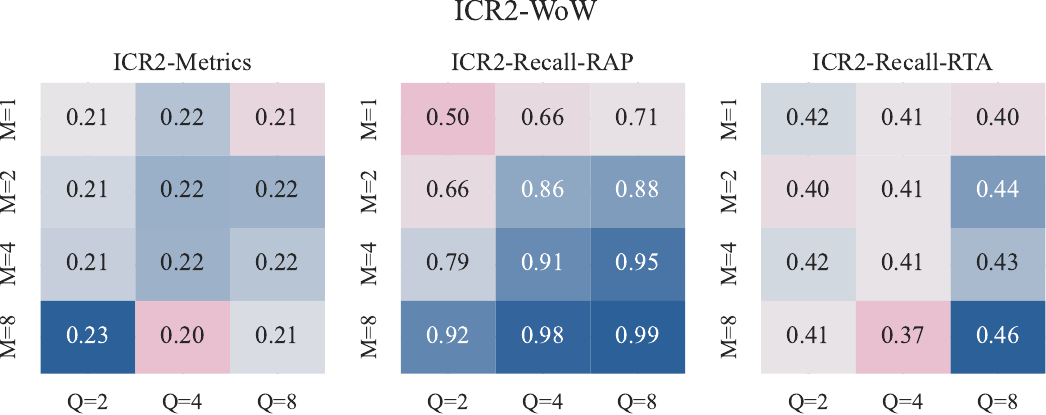}
        \label{fig:subfig7}
    \end{subfigure}
    
    \caption{
Effect of adjusting $Q$ (number of attention heads for retrieval) and $M$ (number of passages to select) when applying SFT-RTA + RAP on all tasks in \icr{}. Left column: average model performance measured by the tasks-specific metrics. Middle column: retrieval performance measured by the recall rate. Right column: retrieval performance measured by the exact match.
    }
    \label{fig:detailed-icr2-rap-results}
\end{figure*}

We show the detailed results for each task of LOFT and \icr{} in Figure~\ref{fig:detailed-loft-rap-results} and~\ref{fig:detailed-icr2-rap-results}.

\section{Details of Experiment Setup}
\label{appendix:detailed-experiment-setup}

\subsection{Training Details}

For all models, we set the base value of the rotary position embedding to $1e6$ following \citet{su2024roformer}. Training is conducted with a batch size of 1 per Nvidia A100-40G GPU, and gradients are accumulated every 4 steps. Each model is trained for up to 10,000 steps (approximately 2 epochs), with the best-performing checkpoint on the validation set selected as part of an early-stopping strategy. The learning rate is set to $1e-5$, and the maximum sequence length during training is capped at 32,768 tokens, discarding any sequences exceeding this threshold.

\subsection{Inference Parameters}

During inference, we use greedy decoding for all models, allowing a maximum generated sequence length of 1,024 tokens. For RAP decoding, 100 random instances from the validation set are used to probe the attention heads responsible for retrieval.
For SFT-DA, retrieval is performed using the designated retrieval attention heads based on the first and only decoded token.
For SFT-RTA, retrieval is performed using all tokens generated in the retrieval step to ensure complete context coverage.

\subsubsection{RAP Hyperparameter Settings}

In this section, we detail the hyperparameters used in applying RAP decoding to SFT-DA and SFT-RTA on \icr{} and LOFT.

For SFT-DA on \icr{}:
\begin{itemize}
    \item \textbf{NaturalQuestions}: $Q=4, M=1$
    \item \textbf{HotpotQA}: $Q=8, M=1$
    \item \textbf{FEVER}: $Q=4, M=4$
    \item \textbf{WoW}: $Q=4, M=4$
\end{itemize}

For SFT-DA on LOFT:
\begin{itemize}
    \item \textbf{NaturalQuestions}: $Q=4, M=4$
    \item \textbf{HotpotQA}: $Q=4, M=4$
    \item \textbf{WoW}: $Q=8, M=2$
\end{itemize}

For SFT-RTA on \icr{}:
\begin{itemize}
    \item \textbf{NaturalQuestions}: $Q=2, M=1$
    \item \textbf{HotpotQA}: $Q=2, M=1$
    \item \textbf{FEVER}: $Q=2, M=8$
    \item \textbf{WoW}: $Q=4, M=8$
\end{itemize}

For SFT-RTA on LOFT:
\begin{itemize}
    \item \textbf{NaturalQuestions}: $Q=8, M=4$
    \item \textbf{HotpotQA}: $Q=4, M=2$
    \item \textbf{WoW}: $Q=8, M=2$
\end{itemize}

\section{Prompt Templates for Supervised Fine-tuning}

\begin{figure*}[h]
    \centering
    \includegraphics[width=0.9\linewidth]{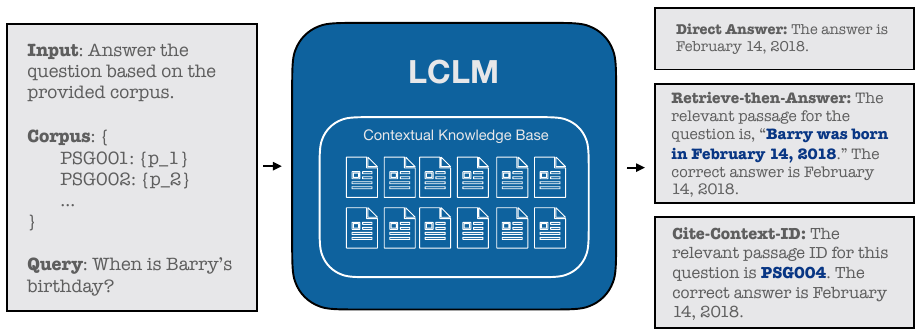}
    \caption{Templates for Direct Answer (DA), Retrieve-Then-Answer (RTA) and Cite-Context-ID (CCI).
    }
    \label{fig:prompt-template}
\end{figure*}

We visualize three templates used for supervised fine-tuning LCLMs to perform retrieve-then-generate generation in Figure~\ref{fig:prompt-template}.

\section{Visualization for Joint Retrieval Head Training}

\begin{figure}[h!]
    \centering
    \includegraphics[width=0.8\linewidth]{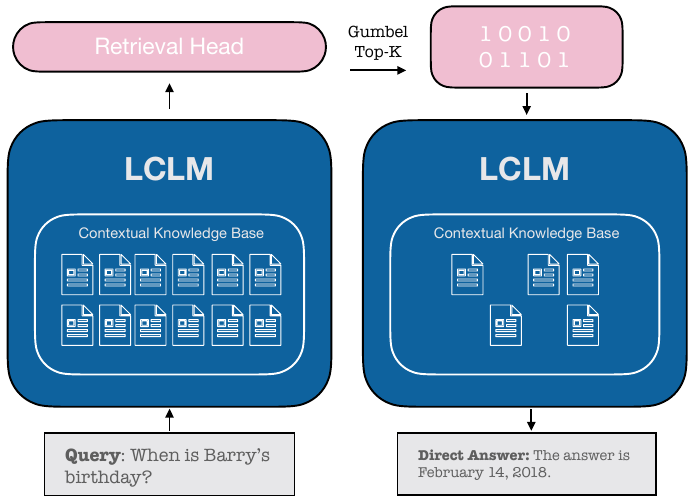}
    \caption{A retrieval head is added to predict the Top-K context for generating final responses. The head is jointly optimized with the generation head. }
    \label{fig:retrieval-head-modeling}
\end{figure}

We visualize the architecture for adding a retrieval head jointly trained with language generation in Figure~\ref{fig:retrieval-head-modeling} for easing the understanding of the approach.

\section{Effect of Retrieval Delineation}

\begin{table}[]
\centering
\resizebox{\linewidth}{!}{%
\begin{tabular}{clrrrr}
\toprule
\multicolumn{2}{c}{\textbf{Benchmark}}          & \cellcolor{gray!20}\textbf{SFT-RTA} & {$w/o$ \texttt{<RET>}} & \cellcolor{gray!20}\textbf{SFT-CCI} & {$w/o$ \texttt{<RET>}} \\ \midrule
\multirow{3}{*}{\rotatebox{90}{\textbf{LOFT}}} & \textbf{NQ}    & \cellcolor{gray!20}0.74                                 & 0.7 \textcolor{myred}{($\downarrow.04$)}                         & \cellcolor{gray!20}0.76                                 & 0.63 \textcolor{myred}{($\downarrow.13$)}                        \\
                               & \textbf{HPQA}  & \cellcolor{gray!20}0.69                                 & 0.66 \textcolor{myred}{($\downarrow.03$)}                        & \cellcolor{gray!20}0.54                                 & 0.58 \textcolor{myblue}{($\uparrow.04$)}                        \\
                               & \textbf{MUSI.} & \cellcolor{gray!20}0.33                                 & 0.24 \textcolor{myred}{($\downarrow.09$)}                        & \cellcolor{gray!20}0.35                                 & 0.24 \textcolor{myred}{($\downarrow.11$)}                        \\ \midrule
\multirow{4}{*}{\rotatebox{90}{\textbf{\icr}}} & \textbf{NQ}    & \cellcolor{gray!20}0.60                                 & 0.56 \textcolor{myred}{($\downarrow.04$)}                        & \cellcolor{gray!20}0.63                                 & 0.61 \textcolor{myred}{($\downarrow.02$)}                        \\
                               & \textbf{HPQA}  & \cellcolor{gray!20}0.67                                 & 0.7 \textcolor{myred}{($\downarrow.03$)}                         & \cellcolor{gray!20}0.63                                 & 0.65 \textcolor{myblue}{($\uparrow.02$)}                        \\
                               & \textbf{FEVER} & \cellcolor{gray!20}0.91                                 & 0.92 \textcolor{myblue}{($\uparrow.01$)}                        & \cellcolor{gray!20}0.89                                 & 0.86 \textcolor{myred}{($\downarrow.03$)}                        \\
                               & \textbf{WoW}   & \cellcolor{gray!20}0.22                                 & 0.20 \textcolor{myred}{($\downarrow.02$)}                        & \cellcolor{gray!20}0.21                                 & 0.22 \textcolor{myblue}{($\uparrow.01$)}   \\ \bottomrule                    
\end{tabular}}
\caption{Effect of removing the special boundary tokens for the retrieve-then-generate variants. }
\label{tab:ablation-study-retrieval-tokens}
\end{table}

As outlined in Section~\ref{sec:retrieve-then-generate-tuning}, our retrieve-then-generate variants, SFT-RTA and SFT-CCI, utilize special symbols \texttt{<RETRIEVAL>} and \texttt{</RETRIEVAL>} to explicitly mark the boundaries of the retrieval step. To examine the impact of these boundary tokens, we conducted an ablation study. As shown in Table~\ref{tab:ablation-study-retrieval-tokens}, removing these tokens in general leads to a decline in model performance. This suggests that clearly delineating the retrieval step is crucial for the model to learn effectively.

\end{document}